\documentclass[10pt,twocolumn,letterpaper]{article}

\usepackage[svgnames]{xcolor}
\usepackage{cvpr}
\usepackage{times}
\usepackage{epsfig}
\usepackage{graphicx}
\usepackage{amsmath}
\usepackage{amssymb}
\usepackage{enumitem}
\usepackage{booktabs}
\usepackage{subfig}
\usepackage[numbers]{natbib}

\usepackage{multirow}
\usepackage{placeins}

\usepackage{array}
\newcolumntype{P}[1]{>{\centering\arraybackslash}p{#1}}

\newcommand{\xikl}{x^i_{k,\ell}}
\newcommand{\xiKL}{x^i_{1:K,1:L}}
\newcommand{\eik}{e^i_k}
\newcommand{\eione}{e^i_1}
\newcommand{\eiK}{e^i_K}
\newcommand{\xikone}{x^i_{k,1}}
\newcommand{\yikL}{y^i_{k,L}}
\newcommand{\yikone}{y^i_{k,1}}
\newcommand{\yikl}{y^i_{k,\ell}}
\newcommand{\xikL}{x^i_{k,L}}
\newcommand{\yim}{y^i_m}
\newcommand{\ls}{L_\text{S}}
\newcommand{\lv}{L_\text{V}}
\newcommand{\Ls}{\mathcal L_\text{S}}
\newcommand{\Lv}{\mathcal L_\text{V}}
\newcommand{\Lssl}{\mathcal L_\text{SSL}}
\newcommand{\Lsup}{\mathcal L_\text{SUP}}
\newcommand{\Lnce}{\mathcal L_\text{NCE}}
\newcommand{\gm}{g_m}

\usepackage[acronym]{glossaries}
\newacronym{vivi}{VIVI}{Video-Induced Visual Invariances}
\newacronym{vtab}{VTAB}{Visual Task Adaptation Benchmark}
\newacronym{aa}{AA}{AutoAugment}

\newacronym{ssl}{SSL}{self-supervised learning}
\newacronym{mt-ssl}{MT-SSL}{multi-task SSL}
\newacronym{ms}{MS}{motion segmentation}
\newacronym{ti}{TI}{transitive invariance}

\newacronym{yt8m}{YT8M}{YouTube-8M}

\newacronym{mlp}{MLP}{multilayer perceptron}
\newacronym{cnn}{CNN}{convolutional neural network}
\newacronym{lstm}{LSTM}{Long Short-Term Memory}
\newacronym{rnn}{RNN}{recurrent neural network}

\newacronym{sgd}{SGD}{stochastic gradient descent}

\newacronym{mse}{MSE}{mean-square error}

\newacronym{nlp}{NLP}{natural language processing}

\usepackage[pagebackref=true,breaklinks=true,colorlinks,bookmarks=false]{hyperref}
\hypersetup{
  colorlinks,
  breaklinks=true,
  anchorcolor={blue!50!black},
  linkcolor={blue!50!black},
  citecolor={blue!50!black},
  urlcolor={blue}
}

\usepackage[group-separator={,},group-minimum-digits=4]{siunitx}
\usepackage{threeparttable}
\usepackage{tabularx}
\usepackage{colortbl}
\usepackage{float}
\usepackage[capitalise]{cleveref}

\newcommand*{\thead}[1]{\multicolumn{1}{c}{#1}}

\renewcommand{\paragraph}[1]{\noindent{\bf #1}\quad}

\usepackage{tikz}
\definecolor{natural}{rgb}{0.7137,0.3333,0.3333}
\definecolor{specialized}{rgb}{0.4118,0.6431,0.4314}
\definecolor{structured}{rgb}{0.3254,0.4431,0.6666}
\definecolor{all}{rgb}{0.7529,0.4902,0.6471}

\cvprfinalcopy 

\def\cvprPaperID{8249} 

\ifcvprfinal\fi
\begin{document}

\title{Self-Supervised Learning of Video-Induced Visual Invariances}

\author{Michael Tschannen \quad Josip Djolonga \quad Marvin Ritter \quad Aravindh Mahendran \\
Xiaohua Zhai \quad Neil Houlsby \quad Sylvain Gelly \quad Mario Lucic \\[0.2cm]
Google Research, Brain Team\vspace{-0.1cm}}

\maketitle
\thispagestyle{empty}
\begin{abstract}
   We propose a general framework for self-supervised learning of transferable visual representations based on \gls{vivi}.
We consider the implicit hierarchy present in the videos and make use of (i) frame-level invariances (e.g.\ stability to color and contrast perturbations), (ii) shot/clip-level invariances (e.g.\ robustness to changes in object orientation and lighting conditions), and (iii) video-level invariances (semantic relationships of scenes across shots/clips), to define a holistic self-supervised loss. Training models using different variants of the proposed framework on videos from the \gls{yt8m} data set, we obtain state-of-the-art self-supervised transfer learning results on the 19 diverse downstream tasks of the \gls{vtab}, using only 1000 labels per task. We then show how to co-train our models jointly with labeled images, outperforming an ImageNet-pretrained ResNet-50 by 0.8 points with 10$\times$ fewer labeled images, as well as the previous best supervised model by 3.7 points using the full ImageNet data set.
\end{abstract}

\section{Introduction}
Supervised deep learning necessitates the collection and manual annotation of large amounts of data, which is often expensive, hard to scale, and may require domain expertise (e.g.,\ in the context of medical data). Expensive data annotation hence presents a bottleneck which impedes the application of deep learning methods to diverse, previously under-explored problems.
Learning \emph{transferable visual representations}, namely representations obtained by training a model on one task (or collection of tasks) which can then be adapted to multiple unseen downstream tasks using \emph{few samples}, is therefore a key research challenge \cite{zhai2019visual}.

An emerging body of work based on \emph{self-supervision} has demonstrated that it is possible to learn such transferable visual representations. The idea is to carefully construct a \emph{pretext} task which does not rely on manual annotation, yet encourages the model to extract useful features from the input. Videos are a promising data modality to design such pretexts tasks for as they capture variations of the instances over time which are not present in images. In addition, there is an abundance of videos available on the Internet covering almost any imaginable domain. 
As a result, and with the recent emergence of research video data sets~\cite{abu2016youtube, thomee2016yfcc100m}, videos have been investigated in the context of self-supervision (for example,~\cite{misra2016shuffle,wei2018learning,wang2017transitive,isola2015learning,wiskott2002slow,zou2011unsupervised,mobahi2009deep,sayed2018cross,ngiam2011multimodal,arandjelovic2017look,agrawal2015learning}). We believe that a holistic approach which captures these diverse efforts can be coupled with image-based pretext tasks to further improve the performance of self-supervised models.

\begin{table}[t]
    \begin{minipage}{\columnwidth}
    \centering
    \small
    \small
\setlength{\tabcolsep}{3pt}
\setlength{\extrarowheight}{5pt}
\renewcommand{\arraystretch}{0.75}
\newcommand{\green}[1]{\textcolor{Green}{(#1)}}
\begin{tabular}{lrlccc}
\toprule
\textsc{Method}                     & \multicolumn{2}{l}{\textsc{Mean}}                          & \textsc{Nat.} & \textsc{Spec.} & \textsc{Str.} \\
\midrule
Ex-ImageNet                         & 59.5          &                          & 50.5          & \bf 81.4       & 56.4 \\
VIVI-Ex(4)                          & 62.5          & \green{+3.0}                   & 55.9          & 80.9           & 59.1 \\
VIVI-Ex(4)-Big                      & \bf 63.3      & \green{+3.8}                   & \bf 57.5      & 81.0           & \bf 59.5 \\
          \midrule
Semi-Ex-10\%~\cite{zhai2019visual}  & 65.3          &                          & \bf 70.2      & 81.9           & 52.7 \\
VIVI-Ex(4)-Co(10\%)                 & \bf 67.2      & \green{+1.9}                   & 63.3          & \bf 82.6       & \bf 62.9 \\
          \midrule
Sup-100\%~\cite{zhai2019visual}     & 66.4          &                          & 73.5          & 82.5           & 52.1 \\
Sup-Rot-100\%~\cite{zhai2019visual} & 68.0          & \textcolor{DimGray}{(+1.6)} & \bf 73.6      & 83.1           & 55.5 \\
VIVI-Ex(4)-Co(100\%)                & 69.4          & \green{+3.0}                   & 69.9          & 83.3           & 62.1 \\
VIVI-Ex(4)-Co(100\%)-Big            & \bf 71.7      & \green{+5.3}                    & 72.5          & \bf 84.3       & \bf 64.7 \\
\bottomrule
\end{tabular}

    \caption[cotrain-caption]{Mean testing accuracy and per-category mean accuracy for models fine-tuned on the 19 diverse downstream tasks (based on \textsc{Nat}ural, \textsc{Spec}ialized, \textsc{Str}uctured data sets) from the \gls{vtab} benchmark \cite{zhai2019visual}\footnotemark, using only 1000 labels per task. The proposed unsupervised models (VIVI-Ex(4) / VIVI-Ex(4)-Big) trained on raw \gls{yt8m} videos and variants co-trained with 10\%/100\% labeled ImageNet data (VIVI-Ex(4)-Co(10\%) / VIVI-Ex(4)-Co(100\%)), outperform the corresponding unsupervised (Ex-ImageNet), semi-supervised (Semi-Ex-10\%) and fully supervised (Sup-100\%, Sup-Rot-100\%) baselines by a large margin.}
    \label{tab:cotraining}
    \end{minipage}
\end{table}
\footnotetext{{\bf We use Version 1 of the \gls{vtab} benchmark} (arXiv:1910.04867v1); please see Appendix~\ref{sec:vtab-v2} for Version 2 (arXiv:1910.04867v2) results.}
In this work we propose a novel, versatile video-based self-supervision framework for learning \emph{image} representations. We divide a video data set into its natural hierarchy of frames, shots, and videos. The intuition is that the model can leverage (1) the \emph{frames} to learn to be robust to color perturbations or contrast changes, (2) the \emph{shot} information to be robust to rigid and non-rigid transformations of objects in a scene, and that (3) explicitly accounting for the \emph{video-level} context should encourage the model to capture semantic relationships of scenes across shots/clips. In contrast to individual frame, shot, or video-level self-supervision objectives, our holistic approach yields a representation that transfers better to a large set of downstream tasks.
As an additional benefit, our approach does not need to pre-compute optical flow or motion segmentation masks, nor does it rely on object tracking. 

We train the proposed model on the \acrfull{yt8m} data set (without using video-level labels) and show that this approach leads to state-of-the-art self-supervised results on the 19 diverse downstream tasks of the \acrfull{vtab} \cite{zhai2019visual}. We then show how to co-train the model jointly with labeled images, outperforming an ImageNet-pretrained ResNet-50 with $10\times$ fewer labeled images. We also investigate the robustness of our co-training models to natural perturbations as induced by the variations across nearby frames in videos \cite{shankar2019systematic}. In summary, our contributions are:
\begin{itemize}[itemsep=2pt,parsep=2pt]
\item We propose a versatile framework to learn image representations from \emph{non-curated videos in the wild} by learning frame-, shot-, and video-level invariances. 
\item We train a variety of models on $3.7$M videos from the \gls{yt8m} data set and achieve a $3.8$\% absolute improvement over image/frame-based baselines across the 19 diverse tasks of the \gls{vtab} benchmark~\cite{zhai2019visual}, which sets new state of the art among unsupervised methods.
\item We augment the \gls{ssl} training framework with a supervised classification loss using data from ImageNet. The resulting models outperform an ImageNet-pretrained network using only 10\% labeled ImageNet images (and no additional unlabeled ones), and achieve a new state of the art when co-trained with the full ImageNet data set, outperforming the best previous supervised result by $3.7$ points.
\end{itemize}

\begin{figure*}[ht!]
\vspace{-0.5cm}
  \centering
\includegraphics[width=.57\textwidth]{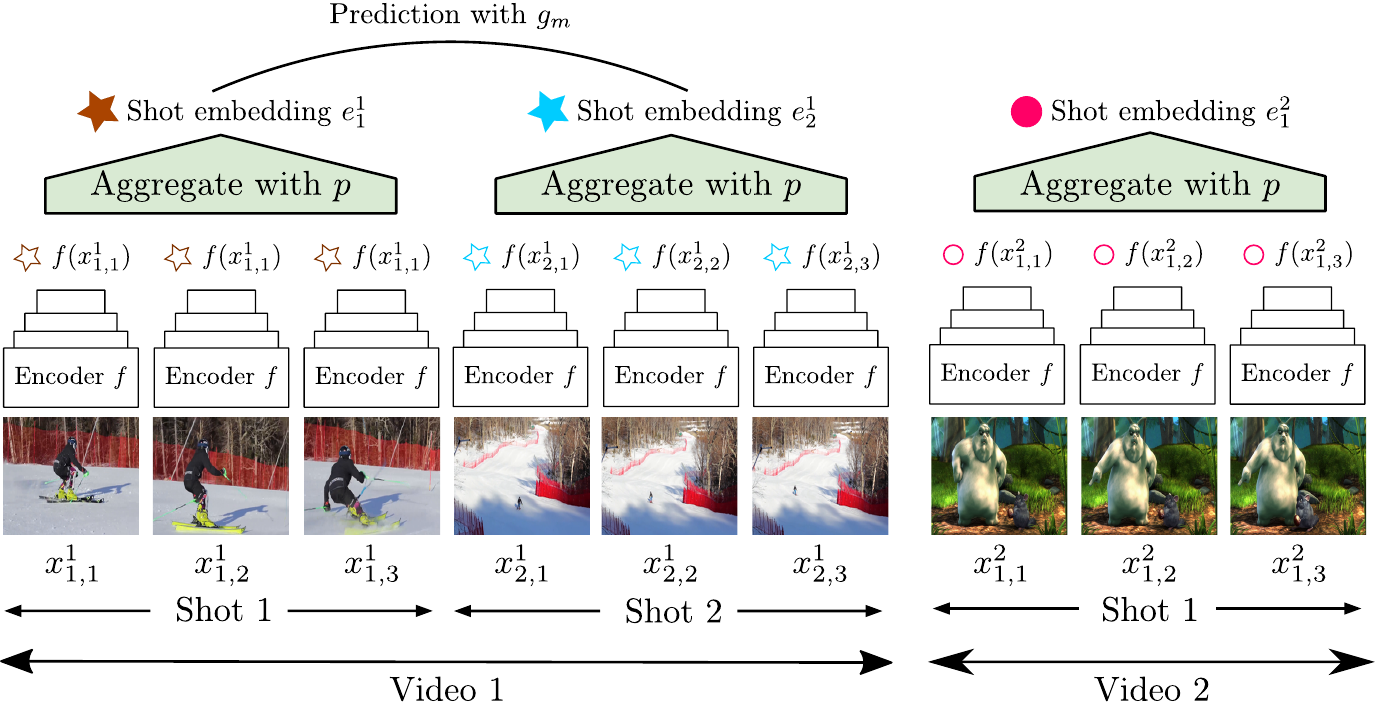}\hspace{5mm}
\includegraphics[width=.35\textwidth]{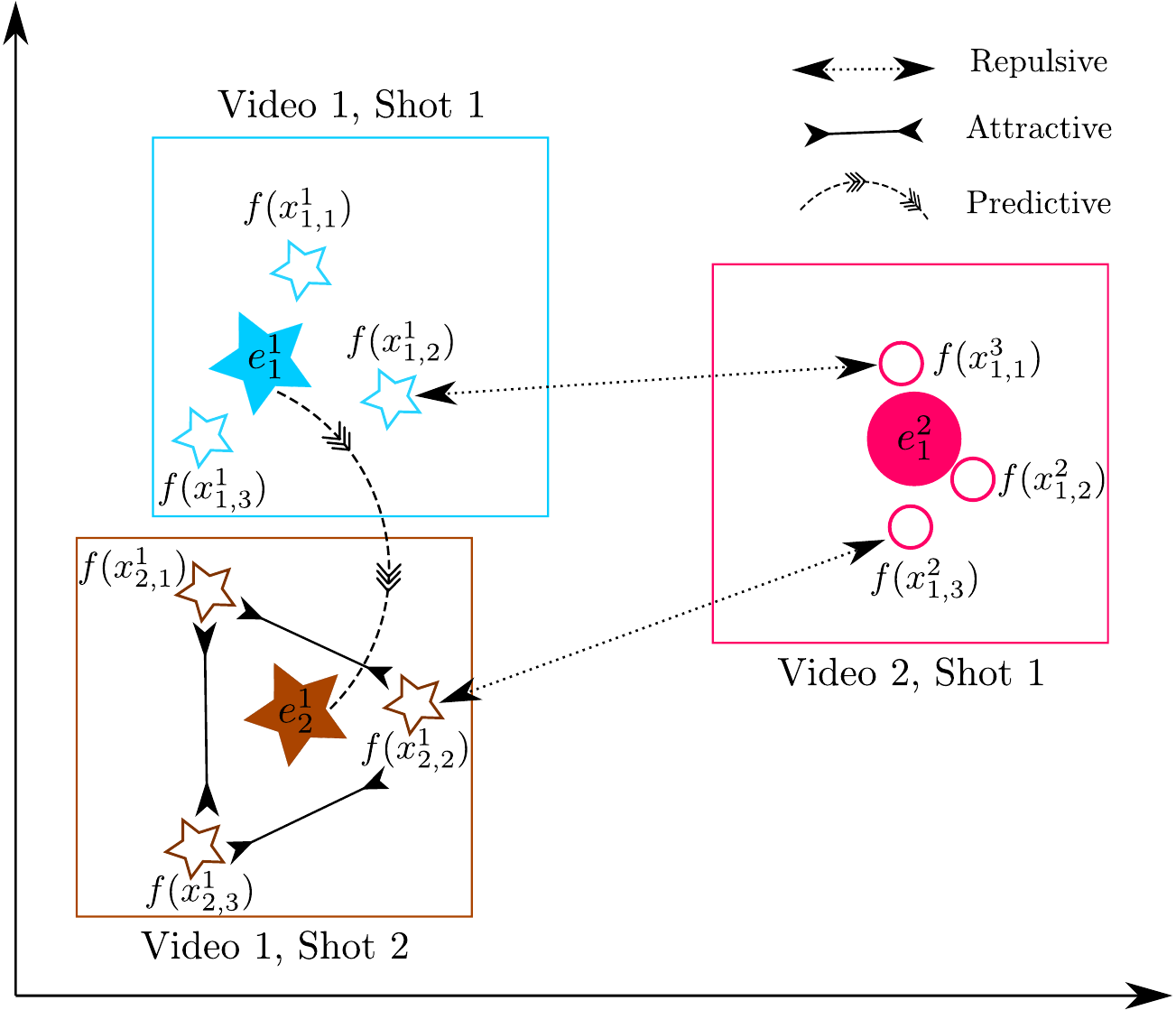}
\caption[Overview]{({\bf left}) Illustration of the frame-, shot-, and video-level encoding pipeline used in this work. Each frame $\xikl$ is encoded using the frame encoder $f$. The frame embeddings $f(\xikl)$  are then aggregated for each shot using a pooling function $p$ to obtain shot embeddings $\eik$. Predictions on the video level are then computed using the prediction functions $g_m$.

({\bf right}) Intuitively, we want to choose frame/shot- and video-level losses that embed frames from the same shot close to each other and frames from different shots or videos far apart, while encouraging shot embeddings from the same video to be predictive of each other using (simple) prediction functions.\footnotemark}
\label{fig:award}
\end{figure*}

\section{Related work}

\paragraph{Self-supervised learning of image representations} 
\gls{ssl} is an active topic of research in the computer vision community. Recent methods \cite{wu2018unsupervised,hjelm2018learning,bachman2019learning,oord2018representation,henaff2019data,tian2019contrastive} have advanced the state of the art in terms of learning representations that can linearly separate between the 1000 ImageNet categories~\cite{imagenet}. Prior work has explored diverse self-supervision cues such as predicting the spatial-context~\cite{doersch2015unsupervised}, colorization~\cite{zhang2016colorful}, equivariance to transformations~\cite{gidaris2018unsupervised,noroozi2017representation}; alongside unsupervised techniques such as clustering~\cite{caron2018deep,zhuang2019local}, generative modelling~\cite{donahue2016adversarial,kingma2014semi}, and exemplar learning~\cite{dosovitskiy2014discriminative}. We adopt some of these \gls{ssl} losses in our framework at the frame-level.

\paragraph{Learning image representations from videos} More relevant to our contribution is the body of literature on \gls{ssl} of image representations from videos. The \emph{temporal context} of frames in video data has been widely exploited.
For example, \cite{misra2016shuffle,lee2017unsupervised,fernando2017self,buchler2018improving,wei2018learning} make use of the order in which frames appear in a video. 
Other forms of temporal context include its combination with spatial context~\cite{wang2017transitive}, and the use of spatio-temporal co-occurrence statistics~\cite{isola2015learning}. Orthogonal to these efforts, which attempt to be selective of the differences between frames, prior work along the lines of \emph{slow feature analysis}~\cite{wiskott2002slow,zou2011unsupervised} also exploited videos as a means of learning invariant representations. Temporal coherence was exploited in a co-training setting by early work~\cite{mobahi2009deep} on learning \glspl{cnn} for visual object recognition and face recognition. Slow and steady feature analysis~\cite{jayaraman2016slow} attempts to learn representations that exhibit higher order temporal coherence. This object deformation signal can be separated from global camera motion by tracking objects using unsupervised methods. These tracked patches have been used to learn image representations~\cite{wang2015unsupervised}. Tracking 
may also be replaced by spatio-temporally matched region proposals~\cite{gao2016object}. Motivated by these works, we explore learning invariances from temporal information in video pixels.

Some of the earliest work making use of temporal consistency used \emph{future frame prediction}~\cite{srivastava2015unsupervised} as a pretext task. A more challenging version of this task is single frame future synthesis. The ambiguity in single-frame prediction has been side-stepped via time-agnostic prediction~\cite{jayaraman2019time}, motion segmentation~\cite{pathak2017learning}, cross-pixel matching~\cite{mahendran2018cross}, and by giving the model a motion cue as input~\cite{zhan2019self}. The latter two require distilling the temporal information from video pixels into optical-flow fields. 

Optical flow has been treated as a separate modality from the RGB pixels in a \emph{multi-modal} setting~\cite{sayed2018cross,tian2019contrastive}. Beyond optical-flow, videos on the web are inherently multi-modal, as they contain audio and subtitles. Multi-modal learning methods that combine vision and audio~\cite{ngiam2011multimodal,de1994learning,owens2016ambient,arandjelovic2017look}, and vision and text~\cite{sun2019videobert} achieve better performance than uni-modal baselines. In a robotics setting, RGB pixels may be considered together with ego-motion~\cite{agrawal2015learning,jayaraman2017learning}. Time-contrastive networks~\cite{sermanet2018time} consider two views of the same action to learn invariant representations. 

Doersch et al.\ \cite{doersch2017multi} show that motion-based \gls{ssl} may be combined with other self-supervision cues namely exemplar, colorization, and spatial-context, to pre-train models that perform better than each of these cues individually. Taking inspiration from their success our framework presents a \emph{synergistic combination} of \gls{ssl} methods.

\paragraph{Transferable representations}
\looseness-1
 Fine-tuning models pre-trained on ImageNet labels is a popular strategy for transferring representations to new tasks~\cite{huh2016makes}.
Kornblith et al.\ \cite{kornblith2018better} show that better supervised models tend to transfer better when fine-tuned.
Other supervised learning benchmarks focus on performance on multiple data sets, either via transfer learning, meta-learning, or multi-task learning~\cite{rebuffi2017,triantafillou2019meta}.
In the representation learning literature, models are usually evaluated in-domain, typically on ImageNet~\cite[and references therein]{zhan2019self}.
However, self-supervised models are now performing well on tasks such as surface normal estimation, detection, and navigation~\cite{goyal2019scaling}.
The \gls{vtab} benchmark evaluates the transferability of representations beyond object classification in the natural image domain to many domains and task semantics such as counting and localization~\cite{zhai2019visual}. 
Similarly, recent developments in \gls{nlp} have lead to representations that transfer effectively to many diverse tasks~\cite{devlin2018bert}.\footnotetext{Video credit: \protect\url{https://vimeo.com/362621732} and \\ \protect\url{https://en.wikipedia.org/wiki/Big_Buck_Bunny}.}

\section{Learning video-induced visual invariances}

We start by giving an overview of the proposed framework in Sec.~\ref{sec:framework-overview}, and discuss frame/shot-level and video-level losses in detail in Sec.~\ref{sec:shot-level-loss} and Sec.~\ref{sec:video-level-loss}, respectively.

\subsection{Overview} \label{sec:framework-overview}

We consider a data set $\mathcal X$ containing $N$ videos, each composed of multiple shots. For simplicity of exposition we assume that each video consists of $K$ shots, and each shot has $L$ frames. If we denote the $\ell$-th frame in the $k$-th shot of video $i$ by $\xikl$, we can write the data set as $\mathcal X = \{\xiKL\}_{i=1}^N$. Our framework consists of a frame-encoder $f$, a frame embedding pooling function $p$, and one or multiple shot-level prediction functions $\gm$ (see Fig.~\ref{fig:award}). The pooling function computes an embedding $\eik$ of the $k$-th shot in video $i$ by feeding each frame through the frame encoder and applying the pooling function,
\begin{equation*}
\eik = p(f(\xikone), \ldots, f(\xikL)).
\end{equation*}
The pooling function can have different forms, ranging from simple average pooling to attention pooling taking the values of the individual frame embeddings $f(\xikl)$ into account. Shot-level prediction functions are trained to predict pretext (label-free) targets from shot embeddings.

We define a frame/shot-level loss and a video-level loss to learn invariances at different levels of abstraction. More specifically, the frame/shot-level loss takes the form
\begin{equation*}
    \Ls = \sum_{i,k} \ls(f(\xikone), \ldots, f(\xikL); \yikone, \ldots, \yikL),
\end{equation*}
where $\yikl$ are shot-level pretext labels and $\ls$ is a shot-level loss that can be instantiated as only acting on the frame level in the sense of $\ls$ decomposing into a sum over the frames $\ell=1,\ldots,L$ (see Sec.~\ref{sec:shot-level-loss} for concrete instantiations of the losses). The video-level loss is given by
\begin{equation} \label{eq:Lv}
    \Lv = \sum_{i,m} \lv(\gm(\eione, \ldots, \eiK)); \yim),
\end{equation}
where the $\yim$ are video-level pretext labels and $\lv$ is a video-level loss (see Sec.~\ref{sec:video-level-loss} for concrete losses). The total loss is then given by $\Lssl = \Ls + \lambda \Lv$, where $\lambda >0$ balances the shot level and video level losses. $\Lssl$ is minimized jointly w.r.t.\ the parameters of $f$, $p$, and $\gm$. 

\paragraph{Co-training with labeled images} We also consider the case where one has access to a limited number of labeled images in addition to the video data. Combining image-based \gls{ssl} losses with a supervised loss applied to a subset of the images was studied previously by \cite{zhai2019s4l}. They found that this approach leads to a state-of-the-art semi-supervised models, and improves the performance of supervised models when all images are labeled. Here, we consider the related setup where the \gls{ssl} loss is computed on video data, and the supervised loss is based on image data from a different data set. Specifically, we additionally apply $f$ followed by a linear classifier to mini-batches of labeled images and compute the cross-entropy loss $\Lsup$ between the predictions and the image labels. The total loss is then computed as $\Lssl + \gamma \Lsup$, where $\gamma > 0$ balances the contributions of the self-supervised and supervised loss terms.

\paragraph{Relation to prior work}
We are not aware of prior work using the natural hierarchy of frame, shot, and video-level invariances in videos for self-supervised image representation learning. Further, our approach is geared towards reducing the need for curated datasets and expensive labeling procedures. In contrast, many existing  methods  for  learning image representations from video data often rely on short curated videos consisting of single clips, or even the treat training set as a bag of frames \cite{caron2019unsupervised, doersch2017multi}.

\subsection{Learning shot-level invariances} \label{sec:shot-level-loss}

To define the frame/shot-level loss $\Ls$, we propose to build on any \gls{ssl} loss designed for images, such as classifying exemplars \cite{dosovitskiy2014discriminative}, solving jigsaw puzzles of image patches \cite{noroozi2016unsupervised}, or rotation prediction \cite{gidaris2018unsupervised}. For learning shot-induced invariances, one can take two approaches:
\begin{enumerate}[label=(\roman*),itemsep=2pt,parsep=2pt]
    \item apply the image-based \gls{ssl} loss independently to each frame so that the shot-induced invariances are learned implicitly through the combination of pooling function and video-level prediction task, or
    \item explicitly ensure that the embeddings of the frames from the same shot are similar by adding a triplet or a contrastive loss to the image-based \gls{ssl} loss.
\end{enumerate}  
In this work, in the spirit of approach (i) we consider \gls{ssl} by rotation prediction \cite{gidaris2018unsupervised} without additional explicit shot-level loss. To explore approach (ii) we rely on a variant of exemplar \gls{ssl}  \cite{dosovitskiy2014discriminative}, where each image is associated with a different class, and a feature extractor is trained to classify each image into its own class after heavily augmenting it (random cropping, rotation, contrast, and color shifts). Following \cite{doersch2015unsupervised, kolesnikov2019revisiting}, to scale this approach to hundreds of millions of images (frames), we employ a triplet loss \cite{schroff2015facenet} encouraging augmentations of the same image to be close and augmentations of different images to be far apart. To learn invariances from different frames of the same shot, rather than picking a random frame from the shot and applying $M$ random augmentations to it, we pick $M$ \emph{consecutive frames} from the same shot and augment each frame once. As a result, our feature extractor learns both the invariances induced by temporal variation in video as well as those induced by the data augmentation.
\subsection{Learning video-level invariances}\label{sec:video-level-loss}

In contrast to action recognition networks, which learn video representations that have to be discriminative w.r.t.\ changes between frames, our framework targets learning representations that are invariant to such changes.
Nevertheless, discriminative tasks useful for learning representations for action recognition, such as predicting whether a sequence of frames is played forward or backward \cite{wei2018learning}, verifying whether the frames are ordered or shuffled \cite{misra2016shuffle}, or predicting features corresponding to future frames \cite{han2019video}, can be useful to learn abstract transferable representations when applied to sensibly chosen \emph{groups of aggregated frames}.
Following this intuition, our framework allows to apply any of these tasks to shot embeddings, rather than individual frame embeddings.
Despite being discriminative at the video level, these tasks encourage the representation to be invariant to all except those cues that are necessary for the pretext task; and hence indirectly induce invariances.
For example, determining whether a sequence of shot embeddings is played forward or backward requires understanding of the high-level semantics of the scene and objects in each shot.
Similarly, predicting future shot embeddings from the past ones encourages learning an abstract summary of each shot.
In this work we will explore these two approaches. 

For shot order prediction, we randomly reverse the order of the shot embeddings and train a prediction function $g$ to predict the shot order from concatenated shot embeddings, i.e., $\lv$ in \eqref{eq:Lv} is the cross-entropy loss and $\yim$ is $1$ if the sequence of shot embeddings is reversed and $0$ otherwise.
To train $g$ to predict future shot embeddings, we rely on noise-contrastive estimation \cite{gutmann2010noise}. Specifically, we use the embeddings of the shots $e^i_1, \ldots, e^i_k$ to obtain a prediction $\hat{e}^i_{k+m}$ of the embedding $e^i_{k+m}$ of the shot $m$ steps in the future.
Then, $\Lv$ should quantify the quality of the prediction, which we accomplish using the InfoNCE loss~\cite{oord2018representation}
\begin{align}\label{eq:infonce}
     \Lnce = -\frac{1}{N}\sum_i \log \frac{e^{g(\hat{e}_{k+m}^i, e_{k+m}^i)}}{\frac{1}{N} \sum_j e^{g(\hat{e}_{k+m}^i, e_{k+m}^j)}},
\end{align}
where $g$ is trained to assign high scores to pairs of shot embeddings from the same video, and low values to embeddings computed from different videos.\footnote{In practice, we use all shot embeddings from the other videos, not only those at time step $k+m$, which is known to improve performance \cite{oord2018representation}.} Note that the terms in \eqref{eq:infonce} can, up to an additive constant, be seen as the cross-entropy loss of an $N$-class classification problem where the correct label is $i$, so that we could reformulate the loss in the form \eqref{eq:Lv} using class labels $y^i$.

\section{Experimental setup} \label{sec:exp-setup}
Our experiments encompass two training phases, which we refer to as \emph{upstream} and \emph{downstream}.
First, in the upstream phase, we train our models on video (and image) data using the methods proposed in the previous section.
Then, we fine-tune those trained models on a set of downstream problems in the second phase.
We focus on the challenging scenario in which the downstream data is limited, and use only 1000 examples for each downstream task~\cite{zhai2019visual}.

\paragraph{Upstream training}
We train on the videos in the \gls{yt8m} data set \cite{abu2016youtube}, which consists of millions of YouTube video IDs with over 3800 visual entities.
We downloaded approximately $4.7$M of these videos sampled at $1$ Hz and split them into a training set of $3.7$M and a testing set of $1$M videos.
We further split the videos into shots using a simple strategy based on color histograms, similarly to \cite{mas2003video} (see Table~\ref{tab:shot-stat} in the supplementary material for data set statistics). No other pre-processing or filtering is performed as we target learning from real-world videos in the wild. We also present results of several baseline approaches applied to a data set obtained by selecting a single random frame from each video, which we refer to as \gls{yt8m} frames.

Furthermore, in the co-training experiments we also use (a class-balanced fraction of) the ImageNet (ILSVRC-2012) training set \cite{deng2009imagenet}. 

\paragraph{Downstream evaluation}
To evaluate the learned representations, we use the data sets and follow the protocol of the {\bf \gls{vtab} Version 1} (arXiv:1910.04867v1) \cite{zhai2019visual}.\footnote{Please see Appendix~\ref{sec:vtab-v2} for Version 2 (arXiv:1910.04867v2) results; the relative improvements of our methods over baselines and the conclusions are similar.}
This protocol consists of 19 tasks categorized into three groups as follows (details and references are in the appendix).
\begin{itemize}[itemsep=2pt,parsep=2pt]
    \item \emph{Natural} --- Six classical image classification problems on natural images (data sets: Caltech101, CIFAR-100, DTD, Flowers102, Pets, Sun397 and SVHN).
    \item \emph{Specialized} --- Image classification on data captured using specialist equipment, from the remote-sensing (data sets: Resisc45, EuroSAT) and medical (data sets: Patch Camelyon, Diabetic Retinopathy) domains.
    \item \emph{Structured} --- Eight tasks to predict properties of the objects appearing in an image (how many there are, their relative position and distance), on both rendered (Clevr, dSprites, SmallNORB, DMLab) and real (KITTI) data.
\end{itemize}

For each of these 19 tasks and each model that we propose, we launch a sweep over 4 hyper-parameters (learning rates and schedules, as in the lightweight mode of \cite{zhai2019visual}).
Then, we choose the models that had the best validation accuracy when averaged over these 19 tasks.
These best-performing models are then re-trained for each data set on 1000 random points from the union of the train and validation set and evaluated on the testing set.
To account for the randomness coming from the initialization of the fresh classification head and the order in which the data appears, we repeated this evaluation scheme three times and report the median test set accuracy (following~\cite{zhai2019visual}).

\begin{figure*}[t!]
\centering
\includegraphics[width=0.98\textwidth]{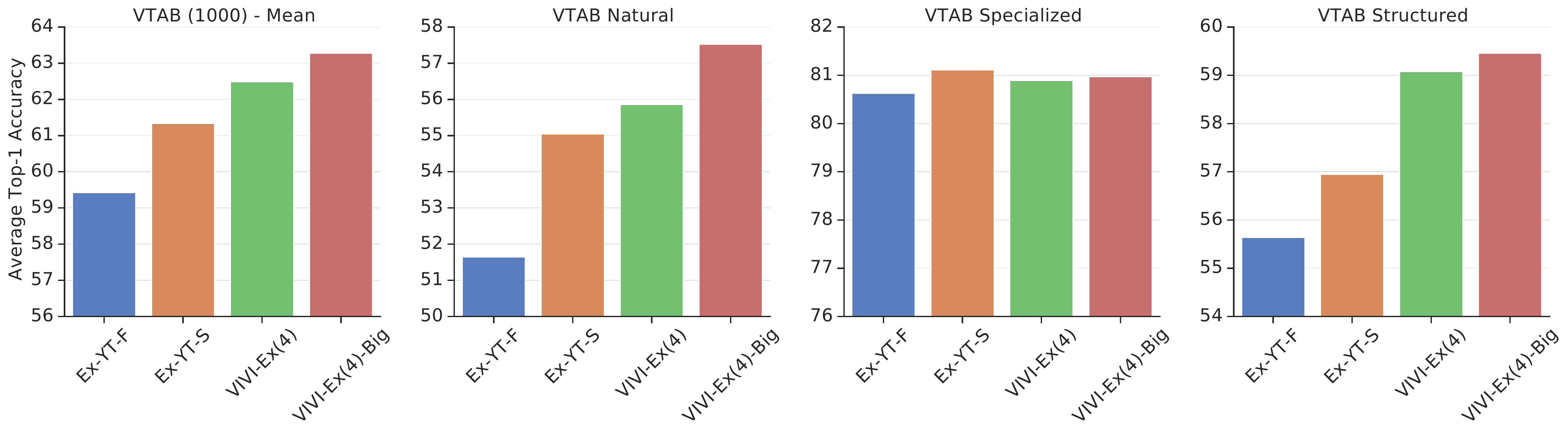}
\vspace{-0.2cm}
\caption{\gls{vtab} 1000 example mean score and per-category mean score of exemplar \gls{ssl} from \gls{yt8m} frames (Ex-YT-F), with additional shot-level self-supervision (Ex-YT-S), the proposed method with InfoNCE video-level prediction across 4 shots (VIVI-Ex(4)) and additionally 3$\times$wider architecture (VIVI-Ex(4)-Big). Both shot and video-level losses improve the overall score, with the gains coming mostly from higher mean accuracy on the natural and structured subsets.}
\vspace{-0.2cm}
\label{fig:pure-ssl}
\end{figure*}

\paragraph{Architectures and training details} The frame encoder $f$ is modeled using the ResNet-50 v2 \cite{he2016identity} architecture with BatchNorm \cite{ioffe2015batch}.
We also investigated the effect of model capacity by widening the network by a factor of three.
To avoid mismatch in batch statistics between the two data sources, in the co-training experiments we replace BatchNorm with GroupNorm \cite{wu2018group} and also standardize \cite{qiao2019weight} the weights of the convolutions.
We construct mini-batches by sampling either 2 or 4 consecutive shots from each video (dropping those videos with fewer shots),
and randomly select 8 consecutive frames for exemplar-based shot-level \gls{ssl} and 4 consecutive frames rotation-based frame-level \gls{ssl}.
For the $\Lnce$ loss, when we sample 2 shots, we predict the embedding of one from the embedding of the other one using a \gls{mlp}, i.e., the function $g$ in \eqref{eq:infonce} has the form $g(e,e')=\phi_1(e)^\top \phi_2(e')$, where $\phi_1, \phi_2$ are \glspl{mlp} with a single hidden layer with 256 units.
In the experiments with 4 shots, we use a \gls{lstm} prediction function with 256 hidden units to predict every shot embedding from the previous ones.
We use temporal order prediction only together with exemplar-based \gls{ssl} and for data with 2 shots per video, relying on a single-hidden-layer \gls{mlp} with 512 hidden units as prediction function.
Throughout, we rely on (parameter-free) average pooling for $p$. For both frame and shot-level \gls{ssl} approaches we use the augmentation mechanism from \cite{szegedy2015going}. For models co-trained with a supervised loss based on a fraction of ImageNet we additionally use the same HSV-space color randomization as \cite{zhai2019s4l}.

We also perform experiments where we replace the augmentation mechanism from \cite{szegedy2015going} with \gls{aa}, which is an augmentation policy learned using a reinforcement learning algorithm from the full ImageNet data set. While this can cause \emph{label leakage} when applied to unsupervised methods, we investigate it to understand how these automatically learned invariances compare to those induced by shot-based augmentation which are label-free.

In all cases we choose the batch size such that the product of the number of videos and the number of shots is 2048, i.e., $NK = 2048$. We train all unsupervised models for 120k iterations, using \gls{sgd} with a learning rate of 0.8 and momentum 0.9, multiplying the learning rate by 0.1 after 90k and 110k iterations. The co-trained models are trained for 100k iterations, and the schedule as well as the batch size is chosen depending on the amount of labeled data used. For the weight $\lambda$ (and $\gamma$ for co-trained models) we sweep over at most four different values. A complete description of all hyper-parameters and architectures can be found in the appendix.

\paragraph{Baselines} We train a rotation and exemplar baseline model on ImageNet and a data set obtained by sampling one frame from each video in our training set (\gls{yt8m} frames). We use the same training protocol as \cite{kolesnikov2019revisiting} for the respective methods except that we increase the batch size to 2048 and stretch the schedule to 120k iterations to be comparable to our methods. Furthermore, for the exemplar-based model we ablate the video-level prediction task, which amounts to treating the shots independently and only using the frames from the same shot as exemplars. In addition, we consider 3 baselines from \cite{zhai2019visual}: A standard ResNet-50 v2 pretrained on ImageNet (achieving top-1/top-5 accuracy of $75.5$\%/$92.6$\% on the ImageNet validation set), the exemplar model trained on ImageNet with 10\% class-balanced labeled data from \cite{zhai2019s4l} (Semi-Ex-10\%), which achieves state-of-the-art semi-supervised accuracy on ImageNet, and the rotation model trained on ImageNet with all labels \cite{zhai2019s4l} (Sup-Rot-100\%).

We further compare against three prior works that learn image representations from video data: The \gls{ms} \cite{pathak2017learning} and the \gls{mt-ssl} models from~\cite{doersch2017multi}, and the \gls{ti} model from~\cite{wang2017transitive}. \gls{ms} learns representations based on a foreground-background segmentation pretext task. The segmentation maps are derived using an off-the-shelf offline video segmentation algorithm. \gls{mt-ssl} combines \gls{ms} and three other self supervision objectives to train a multi-task network.
Its representation derives from a combination of colorization, spatial context, and motion segmentation cues.
The \gls{ms} and \gls{mt-ssl} models fine-tuned in this evaluation have a ResNet-101~\cite{he2016deep} architecture up to the third block.
\gls{ti} builds a graph combining intra-instance and inter-instance edges and exploits transitivity to learn invariant representations. The intra-instance edges are obtained by tracking patches in videos.
We fine-tune their publicly available pre-trained VGG-16~\cite{simonyan2014very} checkpoint.
We refer the reader to the supplementary material for implementation details regarding the evaluation of these baselines.

\section{Results}

In this section we focus on the low sample-size regime, i.e., when each downstream data set consists of 1000 samples, and discuss the performance on the full data sets in the supplementary material (Table~\ref{tab:full-results}). In brief, the ranking of the methods according to the \gls{vtab} mean score using all examples is similar to the ranking according to the \gls{vtab} 1000 example mean score. Further, here we only present the best configuration (w.r.t.\ the number of shots $K$ and choice of prediction function) for each of our \gls{vivi} learning approaches, and defer the results for other configurations to the supplementary material (Table~\ref{tab:full-results}). We also present an evaluation of the proposed methods on object detection in the supplementary material.

\begin{figure*}[t!]
\vspace{-0.2cm}
\centering
\includegraphics[width=0.98\textwidth]{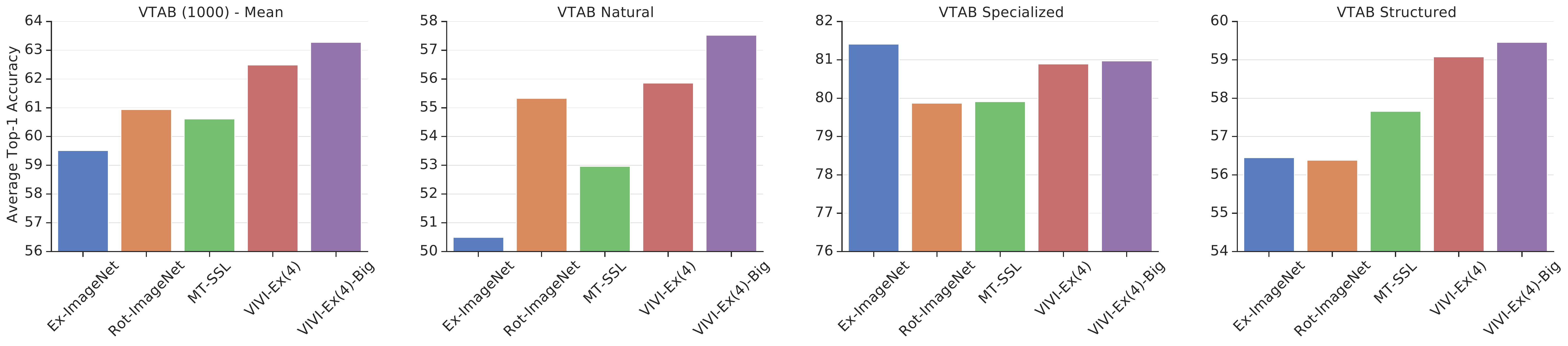}
\vspace{-0.2cm}
\caption{Comparison of the \gls{vtab} 1000 example mean score of the proposed method with exemplar frame/shot-level \gls{ssl} and InfoNCE video-level prediction across 4 shots (VIVI-Ex(4), and with a 3$\times$ wider architecture (VIVI-Ex(4)-Big)), with ImageNet-based exemplar (Ex-ImageNet) and rotation (Rot-ImageNet) baselines, as well as the multi-task \gls{ssl} model from \cite{doersch2017multi}. Our models outperform all baselines on average, and in particular on the structured data sets.}
\label{fig:baselines}
\end{figure*}

\subsection{Self-supervised learning}

\paragraph{Exemplar}
Fig.~\ref{fig:pure-ssl} shows the results for our models and the exemplar-based baselines. The baseline trained on \gls{yt8m} frames only (Ex-YT-F), without leveraging any temporal information, achieves a mean \gls{vtab} 1000 example score of $59.4$\%.
Exploiting the temporal variations within shots to create exemplars (Ex-YT-S) increases that score by about $1.9$ points. Further, adding the video-level prediction loss on top adds another $1.2$ points. It hence appears that leveraging both shot- and video-level invariances using our approach leads to significant gains over just using frames. In addition, increasing the model capacity (using a $3\times$wider model) leads to another increase by $0.8$ points. Note that this model is only $2.0$ points behind the semi-supervised model from \cite{zhai2019s4l} (Semi-Ex-10\%) which uses 128k labeled images from ImageNet for training (cf.\ Table~\ref{tab:cotraining}). The gains mostly come from improvements on the natural and structured data sets, whereas video-level losses do not notably improve the score on the specialized data sets (see Fig.\ \ref{fig:pure-ssl}). We observed the largest gains when using $\Lnce$ with $K=4$ shots and more modest improvements for $\Lnce$ and temporal order prediction with $K=2$ shots.

\vspace{1mm}
\paragraph{Rotation}
Similarly to the exemplar experiments, we observe gains of $2.0$ points in the mean \gls{vtab} 1000 example score over the frame-based baseline (Rot-YT-F) when using a video-level prediction task (VIVI-Rot in Table~\ref{tab:autoaugment}). The gains are smaller for $K=2$ than for $K=4$ shots when combined with $\Lnce$, and temporal order prediction was not effective when combined with rotation prediction as frame-level loss for both $K\in\{2,4\}$. We emphasize that the frame encoder trained via rotation \gls{ssl} on \gls{yt8m} frames performs considerably worse than the same model trained on ImageNet. This is not surprising as ImageNet images are carefully cropped and the data has a balanced class distribution. By contrast, frames sampled from \gls{yt8m} are less balanced in terms of content and arguably provide many shortcuts for the rotation task such as black borders, overlaid logos, frames with text, or lack of orientation cues.

\paragraph{Effect of AutoAugment (AA)} Table~\ref{tab:autoaugment} shows the effect of using \gls{aa}~\cite{cubuk2018autoaugment} instead of the augmentation mechanism from \cite{szegedy2015going}. The effect is strongest on the frame-based baselines, increasing the \gls{vtab} 1000-example score by at least 2, and weakest on models involving shot- and video-level losses, where the increase is between $0.5$ and $1.5$ points. Hence, the invariances induced by \gls{aa} are, to some degree, complementary to the proposed shot- and video-level losses. However, note that \gls{aa} is trained on labeled ImageNet images, which might introduce label leakage.
Hence, methods relying on \gls{aa} should not be considered fully unsupervised.

\begin{table}
    \centering
    \small
    \small
\setlength{\tabcolsep}{4pt}
\setlength{\extrarowheight}{5pt}
\renewcommand{\arraystretch}{0.75}
\begin{tabular}{ccccccc}
\toprule
\multicolumn{5}{c}{\textsc{Exemplar}} & \multicolumn{2}{c}{\textsc{Rotation}}  \\
\cmidrule(r){2-5}
\cmidrule(r){6-7}
& \textsc{yt-f} & \textsc{yt-s} & \textsc{vivi(4)} & \textsc{vivi(4)-Big} & \textsc{yt-f} & \textsc{vivi}  \\
\midrule
\textsc{w/o aa} &     59.4 &     61.3 &       62.5 &           63.3 &     56.9 &        58.9 \\
\textsc{aa}  & 61.8 & 62.8 &       63.0 &           64.4 &     58.9 &    59.9 \\
\bottomrule
\end{tabular}
    \caption{Effect of replacing the data augmentation mechanism from \cite{szegedy2015going} with \gls{aa}. Video-induced invariances learned by our method are complementary to AA in the sense that applying AA to different variants of our method consistently leads to improvements.}\vspace{-4mm}
    \label{tab:autoaugment}
\end{table}

\paragraph{Comparison with related work} Fig.\ \ref{fig:baselines} presents a summary of the comparison with baselines. We omit MS and \gls{ti} as they obtain a \gls{vtab} 1000 example mean score comparable to relative patch location prediction \cite{doersch2015unsupervised} and jigsaw \cite{noroozi2016unsupervised} \gls{ssl} trained on ImageNet. These two methods have a significantly lower \gls{vtab} 1000 example score than the \gls{mt-ssl} model, as well as rotation and exemplar \gls{ssl}.
Our VIVI models clearly outperform both the ImageNet baseline and the \gls{mt-ssl} model. The score obtained by \gls{mt-ssl} is comparable to that obtained by rotation-based \gls{ssl} trained on ImageNet, which in turn scores $1.4$ points higher than exemplar-based SSL. Both our models and \gls{mt-ssl} significantly outperform rotation and exemplar-based \gls{ssl} on the structured data sets, whereas the ImageNet-based exemplar baseline obtains the highest mean score on the specialized data sets.

\subsection{Co-training with ImageNet}
\begin{figure}[t!]
\vspace{-0.2cm}
\centering
\includegraphics[width=0.47\textwidth]{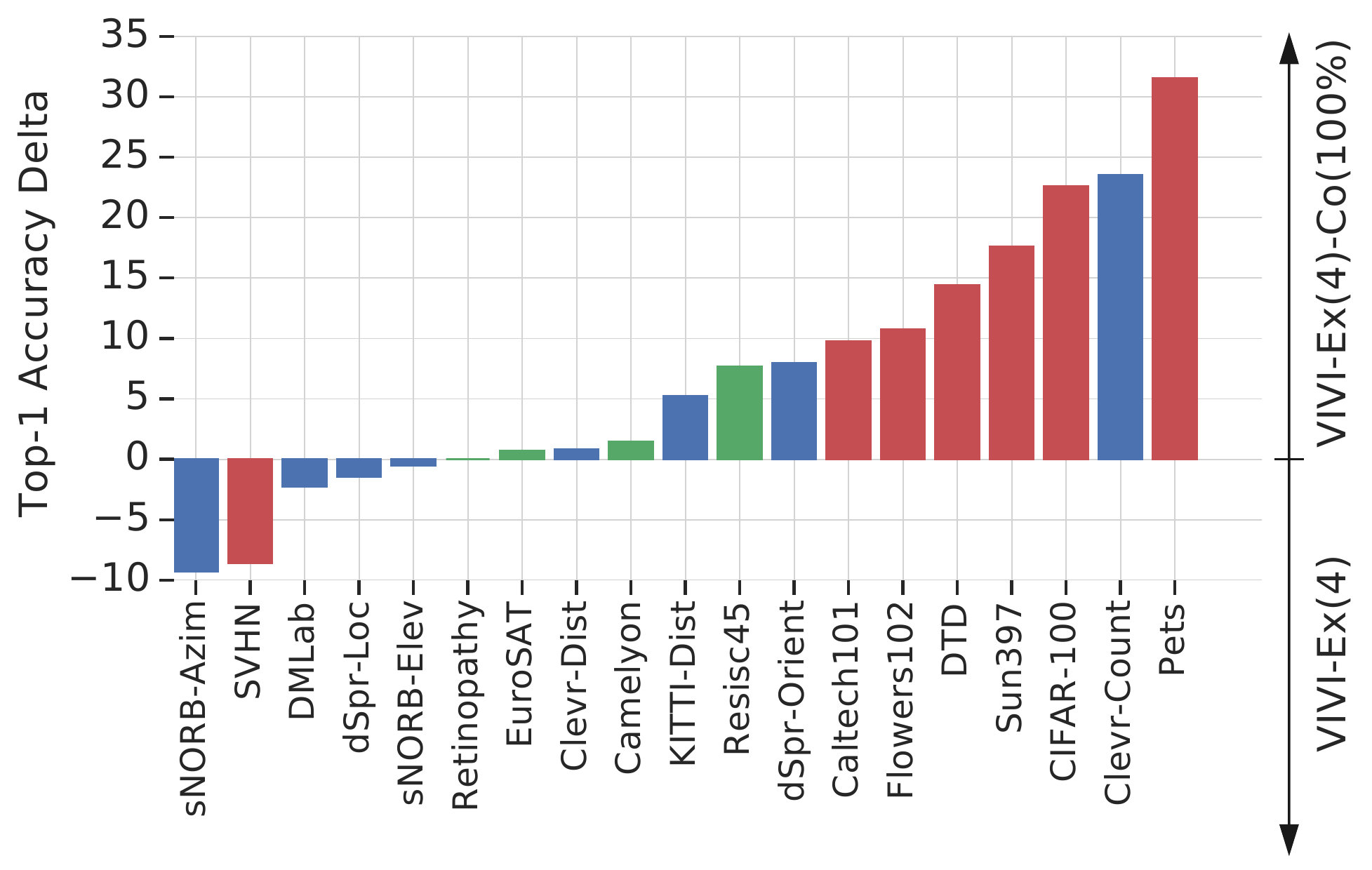}
\vspace{-0.3cm}
\caption{Per-data set comparison of our exemplar-based unsupervised model (VIVI-Ex(4)) and its counterpart co-trained with the full ImageNet data set (VIVI-Ex(4)-Co(100\%)). The accuracy on most of the natural (red) and specialized (green) data sets improves, with the largest improvements observed on the latter, while the accuracy decreases for about half of the structured data sets (blue).}
\label{fig:atari1}
\vspace{-0.1cm}
\end{figure}
In Table~\ref{tab:cotraining} we compare the scores obtained by our exemplar-based co-training models with the baselines from \cite{zhai2019visual}. Our model with frame/shot-level and video-level losses and a wider architecture (VIVI-Ex(4)-Big) reduces the gap between exemplar trained on ImageNet and the strong Semi-Ex-10\% semi-supervised baseline model by more than a factor of 2. Moreover, our model co-trained with 10\% labeled ImageNet examples (class-balanced, no additional unlabeled ImageNet examples are used) outperforms both the Semi-Ex-10\% baseline and the ImageNet pre-trained ResNet-50 on the \gls{vtab} 1000 examples mean score. Using the entire labeled ImageNet training set for co-training yields an increase of $2.1$ points. Finally, scaling up the architecture and applying \gls{aa} to pre-process the ImageNet data adds $2.3$ points, leading to a clear new state of the art on the \gls{vtab} benchmark. The largest gains from using (a subset of) ImageNet can generally be observed on the natural data sets, whereas the gains on the specialized and structured data sets are significantly lower. This result is not surprising given that many data sets in the natural category are semantically similar to ImageNet. Fig.~\ref{fig:atari1} shows the per-data set increase/decrease in the \gls{vtab} 1000 example score when adding a classification loss computed on the entire ImageNet data set to VIVI-Ex(4).

\vspace{1mm}
\paragraph{Robustness to video perturbations} Our co-trained models are trained to both recognize 1000 ImageNet categories and be invariant to deformations found in video data. We therefore expect model predictions to be stable across neighbouring frames in a video.
To measure if this is indeed the case, we evaluate our VIVI-Ex(4)-Co(100\%) model on the \emph{ImageNet-Vid-Robust}~\cite{shankar2019systematic} benchmark.
This benchmark measures the drop in accuracy under a stricter definition of the 0-1 loss using videos from the ImageNet-Vid data set~\cite{imagenet}. Given a set of frames, the prediction on an ``anchor'' frame is considered correct only if \emph{all} neighboring frames are predicted correctly. Intuitively, the drop in performance going from standard top-1 accuracy on anchor frames to this stricter loss function is indicative of a lack in model robustness. The lower the drop the more robust the model. In Table~\ref{cvpr2020:table:imnetvidrobust} we observe that our co-trained model is slightly more robust than its purely supervised counterpart, although the results are still within error bars. This is similar to the difference in performance drop observed for fine-tuning on ImageNet-Vid as reported in the benchmark paper itself~\cite[Table 1]{shankar2019systematic}. These initial results suggest that our co-training approach leads to a similar effect as fine-tuning, despite the domain shift between \gls{yt8m} and ImageNet-Vid.
It seems that robustness to natural perturbations in videos is extremely challenging and worth investigating in the future.


\begin{table}
\small
\centering
\begin{tabular}{lccc}
    \toprule
    Model Type  &  \thead{\shortstack{Accuracy \\ Original}} &  \thead{\shortstack{Accuracy\\ Perturbed}} & $\Delta$ \\
    \midrule
    ImageNet & 68.0 {\footnotesize \textcolor{gray}{[65.2, 70.7]}} & 49.9 {\footnotesize \textcolor{gray}{[46.9, 52.9]}} & 18.1\\
    {\scriptsize VIVI-Ex(4)-Co(100\%)} & 62.2 {\footnotesize \textcolor{gray}{[59.3, 65.1]}} & 46.3 {\footnotesize \textcolor{gray}{[43.3, 49.2]}} & 15.9\\
    \bottomrule
\end{tabular}
\caption{ImageNet-Vid-Robust evaluation: We evaluate our VIVI-Ex(4)-Co(100\%) model (co-trained using all labeled images available in the ImageNet training set), on the ImageNet-Vid-Robust benchmark~\cite{shankar2019systematic}. \emph{Accuracy original} is the top-1 accuracy measured on ``anchor'' frames. \emph{Accuracy perturbed} is the PM-10 accuracy from the benchmark. It is the worst case accuracy defined over neighbouring 20 frames~\cite{shankar2019systematic} around each ``anchor'' frame. $\Delta$ is the absolute difference between these two. On this benchmark, lower difference is better. The small text in gray corresponds to the Clopper-Pearson confidence interval.}
\label{cvpr2020:table:imnetvidrobust}
\end{table}



\section{Conclusion}

We propose and evaluate a versatile framework for learning transferable, data-efficient image representations by exploiting video-induced visual invariances at different levels of granularity. The framework can be instantiated with any image-based \gls{ssl} loss at the frame/shot-level and arbitrary sequence prediction proxy tasks at the video-level. Our experiments reveal that purely self-supervised models benefit greatly from exploiting video-induced invariances, outperforming the \gls{ssl} baselines trained on ImageNet by a large margin, in particular on problems that require predicting the structural properties of the data. Moreover, when augmenting the proposed framework with a supervised classification loss, the resulting models outperform a standard ImageNet-pretrained model using $10\times$ fewer labeled examples, and set a new state of the art on the \gls{vtab} benchmark when co-trained with the full ImageNet data set.

Future research could target better understanding of how the choice of losses and data sets used for upstream training impacts the performance on different tasks in downstream evaluation. While we found our co-trained models to be somewhat more robust to natural perturbations induced by videos than models trained only on images, further research is needed on learning models that overcome robustness issues related to perturbations induced by videos.

\vspace{.1cm}
\paragraph{Acknowledgments} We would like to thank Raphael Marinier for help with preparing the \gls{yt8m} data set. Further, we are grateful to Lucas Beyer for his implementation of GroupNorm with weight standardization.

{
\small
\bibliographystyle{ieee_fullname}
\bibliography{video_refs}
}

\clearpage

\onecolumn

\appendix

\begin{table*}[h!]
    \centering
    \small
    \newcommand{\rotyt}{\rowcolor{orange!20!white}}
\newcommand{\rotytaa}{\rowcolor{orange!40!white}}
\newcommand{\exyt}{\rowcolor{violet!20!white}}
\newcommand{\exytaa}{\rowcolor{violet!40!white}}
\newcommand{\excoyt}{\rowcolor{yellow!20!white}}
\newcommand{\excoytaa}{\rowcolor{yellow!40!white}}
\fontsize{7pt}{7pt}\selectfont
\newcolumntype{C}{>{\centering\arraybackslash}X}
\setlength{\tabcolsep}{0pt}
\setlength{\extrarowheight}{5pt}
\renewcommand{\arraystretch}{0.80}
\begin{tabularx}{\linewidth}{p{10pt}p{2.8cm}!{\color{lightgray}\vline} CCCCCCC!{\color{lightgray}\vline}CCCC!{\color{lightgray}\vline}CCCCCCCC!{\color{lightgray}\vline}C}
\toprule
 &
 & \rotatebox{90}{\tikz\fill[natural] (0,0) circle (.5ex); Caltech101}
 & \rotatebox{90}{\tikz\fill[natural] (0,0) circle (.5ex); CIFAR-100}
 & \rotatebox{90}{\tikz\fill[natural] (0,0) circle (.5ex); DTD}
 & \rotatebox{90}{\tikz\fill[natural] (0,0) circle (.5ex); Flowers102}
 & \rotatebox{90}{\tikz\fill[natural] (0,0) circle (.5ex); Pets}
 & \rotatebox{90}{\tikz\fill[natural] (0,0) circle (.5ex); SVHN}
 & \rotatebox{90}{\tikz\fill[natural] (0,0) circle (.5ex); Sun397}
 & \rotatebox{90}{\tikz\fill[specialized] (0,0) circle (.5ex); Camelyon}
 & \rotatebox{90}{\tikz\fill[specialized] (0,0) circle (.5ex); EuroSAT}
 & \rotatebox{90}{\tikz\fill[specialized] (0,0) circle (.5ex); Resisc45}
 & \rotatebox{90}{\tikz\fill[specialized] (0,0) circle (.5ex); Retinopathy}
 & \rotatebox{90}{\tikz\fill[structured] (0,0) circle (.5ex); Clevr-Count}
 & \rotatebox{90}{\tikz\fill[structured] (0,0) circle (.5ex); Clevr-Dist}
 & \rotatebox{90}{\tikz\fill[structured] (0,0) circle (.5ex); DM-Lab}
 & \rotatebox{90}{\tikz\fill[structured] (0,0) circle (.5ex); KITTI-Dist} 
 & \rotatebox{90}{\tikz\fill[structured] (0,0) circle (.5ex); dSpr-Loc}
 & \rotatebox{90}{\tikz\fill[structured] (0,0) circle (.5ex); dSpr-Ori}
 & \rotatebox{90}{\tikz\fill[structured] (0,0) circle (.5ex); sNORB-Azim}
 & \rotatebox{90}{\tikz\fill[structured] (0,0) circle (.5ex); sNORB-Elev}
 & \rotatebox{90}{\tikz\fill[all] (0,0) circle (.5ex); Mean} \\
 
\midrule

& MS                      &       50.4 &      17.2 & 34.9 &       34.7 & 18.7 & 80.7 &    7.0 &     79.7 &    90.4 &     45.5 &        73.6 &        45.0 &       56.9 &  34.8 &       60.6 &     77.8 &        46.6 &       48.6 &       35.3 & 49.4 \\
& TI                      &       51.6 &      13.4 & 37.7 &       15.5 & 31.1 & 78.8 &    7.2 &     83.2 &    85.7 &     33.6 &        74.3 &        61.2 &       63.9 &  33.1 &       61.6 &     97.4 &        60.4 &       36.5 &       26.1 & 50.1 \\
& Jigsaw                  &       66.7 &      18.9 & 51.4 &       66.1 & 37.5 & 55.1 &   12.1 &     76.0 &    91.5 &     66.2 &        72.4 &        42.8 &       55.9 &  30.5 &       68.2 &     69.5 &        35.0 &       44.9 &       36.3 & 52.5 \\
& Rel.Pat.Loc             &       68.5 &      19.1 & 52.2 &       69.0 & 41.3 & 60.9 &   11.1 &     77.5 &    92.6 &     65.4 &        70.7 &        43.5 &       59.6 &  33.6 &       68.2 &     70.7 &        29.3 &       47.2 &       35.2 & 53.5 \\
\rotyt \cellcolor{white}
& Rot-YT-F                &       70.2 &      21.5 & 48.0 &       48.0 & 35.7 & 88.3 &    8.4 &     83.4 &    93.0 &     61.7 &        73.6 &        48.1 &       57.9 &  39.3 &       73.4 &     90.1 &        51.2 &       50.5 &       38.3 & 56.9 \\
\rotyt \cellcolor{white}
& VIVI-Rot(2)             &       73.8 &      27.0 & 51.2 &       54.4 & 35.7 & 88.2 &   11.6 &     77.7 &    93.1 &     68.6 &        73.6 &        44.0 &       58.0 &  39.1 &       72.5 &     80.0 &        49.8 &       53.0 &       38.8 & 57.4 \\
\rotyt \cellcolor{white}
& VIVI-Rot(4)             &       73.4 &      25.8 & 52.0 &       53.1 & 42.2 & 88.5 &   10.3 &     84.1 &    93.7 &     66.3 &        73.6 &        49.9 &       58.3 &  38.7 &       73.6 &     88.9 &        52.4 &       52.8 &       40.6 & 58.9 \\
\rotytaa \cellcolor{white}
& Rot-YT-F-AA             &       77.8 &      24.0 & 51.2 &       56.1 & 33.3 & 89.1 &   10.0 &     85.1 &    93.9 &     66.5 &        73.6 &        48.8 &       59.2 &  39.0 &       71.2 &     92.4 &        55.1 &       54.6 &       38.7 & 58.9 \\
\exyt \cellcolor{white}
& Ex-YT-F                 &       73.0 &      24.3 & 49.8 &       64.6 & 48.1 & 88.4 &   13.4 &     83.0 &    95.5 &     70.4 &        73.6 &        50.4 &       59.0 &  38.6 &       71.0 &     90.8 &        46.9 &       43.5 &       44.7 & 59.4 \\
& Ex-ImageNet             &       72.0 &      20.1 & 52.8 &       54.5 & 51.0 & 87.5 &   15.5 &     83.8 &    95.2 &     72.5 &        74.2 &        49.9 &       60.9 &  36.9 &       75.6 &     92.2 &        45.6 &       48.8 &       41.7 & 59.5 \\
& MT-SSL                  &       76.1 &      27.1 & 52.9 &       63.2 & 48.2 & 89.6 &   13.6 &     81.5 &    93.5 &     71.0 &        73.6 &        56.6 &       59.4 &  37.3 &       72.9 &     94.7 &        47.4 &       52.3 &       40.7 & 60.6 \\
& Rot-ImageNet            &       80.6 &      25.9 & 56.5 &       72.6 & 47.1 & 88.6 &   16.0 &     81.9 &    94.2 &     69.8 &        73.6 &        49.1 &       58.6 &  37.9 &       73.1 &     92.6 &        50.4 &       51.6 &       37.7 & 60.9 \\
\exyt \cellcolor{white}
& Ex-YT-S                 &       76.2 &      28.4 & 50.4 &       74.9 & 53.1 & 88.0 &   14.3 &     81.7 &    94.8 &     74.2 &        73.7 &        52.8 &       58.9 &  39.3 &       70.9 &     91.1 &        50.8 &       49.7 &       42.1 & 61.3 \\
\exytaa \cellcolor{white}
& Ex-YT-F-AA              &       78.7 &      28.1 & 54.6 &       64.7 & 52.7 & 89.0 &   16.5 &     83.5 &    95.5 &     73.1 &        73.6 &        52.2 &       60.3 &  39.0 &       74.4 &     93.4 &        54.6 &       45.9 &       44.5 & 61.8 \\
\exyt \cellcolor{white}
& VIVI-Ex(2)-Ord          &       76.0 &      29.0 & 49.0 &       77.7 & 54.7 & 88.5 &   13.6 &     80.5 &    94.2 &     73.2 &        73.6 &        55.9 &       60.2 &  39.1 &       72.0 &     91.5 &        52.1 &       51.5 &       42.8 & 61.9 \\
\exyt \cellcolor{white}
& VIVI-Ex(2)              &       75.3 &      28.8 & 48.7 &       77.5 & 55.5 & 87.9 &   12.4 &     81.6 &    94.1 &     73.6 &        73.6 &        56.3 &       60.6 &  38.9 &       73.1 &     91.9 &        52.2 &       50.6 &       44.6 & 62.0 \\
\exyt \cellcolor{white}
& VIVI-Ex(4)              &       76.3 &      29.0 & 50.1 &       77.9 & 55.6 & 88.0 &   14.1 &     82.4 &    94.4 &     73.1 &        73.6 &        55.3 &       60.9 &  38.6 &       72.9 &     95.3 &        53.0 &       52.4 &       44.1 & 62.5 \\
\exytaa \cellcolor{white}
& Ex-YT-S-AA              &       79.3 &      30.1 & 53.9 &       75.4 & 55.3 & 88.4 &   14.7 &     83.4 &    94.8 &     75.7 &        73.6 &        55.3 &       59.5 &  40.7 &       76.6 &     91.3 &        52.9 &       51.2 &       41.4 & 62.8 \\
\exytaa \cellcolor{white}
& VIVI-Ex(4)-AA           &       78.6 &      30.3 & 51.5 &       75.0 & 56.1 & 88.6 &   14.4 &     83.0 &    94.7 &     75.2 &        73.6 &        56.3 &       60.6 &  41.6 &       74.2 &     94.6 &        55.5 &       52.3 &       41.0 & 63.0 \\
\exyt \cellcolor{white} & VIVI-Ex(4)-Big          &       77.5 &      32.8 & 51.3 &       79.4 & 56.6 & 88.3 &   16.6 &     79.8 &    95.1 &     75.3 &        73.6 &        54.7 &       57.9 &  40.4 &       74.4 &     92.0 &        56.8 &       52.4 &       47.0 & 63.3 \\
\exytaa \cellcolor{white}
& VIVI-Ex(4)-Big-AA       &       77.5 &      34.8 & 54.2 &       76.9 & 59.5 & 89.7 &   16.2 &     84.3 &    94.8 &     77.2 &        73.6 &        53.3 &       60.7 &  40.5 &       78.0 &     93.4 &        59.2 &       52.9 &       47.0 & 64.4 \\
& Semi-Ex-10\%             &       88.6 &      53.2 & 60.8 &       86.8 & 85.3 & 88.0 &   29.0 &     83.2 &    95.2 &     77.3 &        71.7 &        42.3 &       57.4 &  36.7 &       71.4 &     74.9 &        53.9 &       52.7 &       32.3 & 65.3 \\
& Sup-100\%                &       91.0 &      57.0 & 66.0 &       88.6 & 89.9 & 87.3 &   34.4 &     80.6 &    95.3 &     80.8 &        73.2 &        41.0 &       56.1 &  36.3 &       70.6 &     85.7 &        46.0 &       45.7 &       35.4 & 66.4 \\
\excoyt \cellcolor{white}
& VIVI-Ex(4)-Co(10\%)      &       82.8 &      36.6 & 58.1 &       82.7 & 76.9 & 81.9 &   24.1 &     85.6 &    94.7 &     76.4 &        73.6 &        79.4 &       63.9 &  38.0 &       76.6 &     95.3 &        61.3 &       42.4 &       46.3 & 67.2 \\
& Sup-Rot-100\%            &       91.7 &      53.7 & 69.5 &       90.8 & 88.1 & 88.5 &   32.8 &     83.4 &    96.0 &     82.0 &        71.1 &        47.3 &       57.2 &  36.6 &       77.1 &     88.3 &        52.1 &       51.6 &       33.7 & 68.0 \\
\excoyt \cellcolor{white}
& VIVI-Ex(4)-Co(100\%)     &       86.1 &      51.5 & 64.5 &       88.7 & 87.1 & 79.4 &   31.7 &     83.9 &    95.1 &     80.8 &        73.6 &        78.9 &       61.7 &  36.4 &       78.2 &     93.8 &        61.0 &       43.1 &       43.6 & 69.4 \\
\excoytaa \cellcolor{white}
\multirow{-27}{*}{\rotatebox{90}{1000}}& VIVI-Ex(4)-Co(100\%)-Big &       88.0 &      53.3 & 69.0 &       90.4 & 88.4 & 84.4 &   34.1 &     86.2 &    95.9 &     81.7 &        73.6 &        79.9 &       63.5 &  37.3 &   82.9 &     95.3 &        67.4 &       46.2 &       44.9 & 71.7 \\

\arrayrulecolor{lightgray}\specialrule{.5pt}{0.6pt}{-0.5pt}\arrayrulecolor{black}

& MS                      &       68.4 &      69.6 & 48.1 &       52.7 & 49.2 & 96.7 &   56.9 &     85.5 &    97.5 &     88.3 &        76.8 &        99.8 &       90.4 &  71.7 &       75.3 &    100.0 &        96.3 &       99.9 &       97.4 & 80.0 \\
& TI                      &       76.5 &      68.5 & 56.4 &       66.3 & 52.0 & 96.2 &   59.4 &     89.8 &    97.6 &     90.1 &        81.0 &        94.0 &       91.6 &  72.3 &       61.2 &    100.0 &        96.4 &       97.0 &       86.2 & 80.7 \\
& Jigsaw                  &       79.1 &      65.3 & 63.9 &       77.9 & 65.4 & 93.9 &   59.2 &     83.0 &    97.9 &     92.0 &        80.1 &        99.6 &       88.6 &  72.0 &       74.7 &    100.0 &        90.3 &       99.9 &       93.6 & 83.0 \\
& Rel.Pat.Loc             &       79.9 &      65.7 & 65.2 &       78.8 & 66.8 & 93.7 &   58.0 &     85.3 &    97.8 &     91.5 &        79.8 &        99.5 &       87.7 &  71.5 &       75.0 &    100.0 &        90.4 &       99.7 &       92.6 & 83.1 \\
\rotyt \cellcolor{white}
& Rot-YT-F                &       81.8 &      72.6 & 60.7 &       66.5 & 65.7 & 96.9 &   59.4 &     86.7 &    98.3 &     92.2 &        76.8 &        99.8 &       92.1 &  76.0 &       81.3 &    100.0 &        96.6 &       99.8 &       98.0 & 84.3 \\
\rotyt \cellcolor{white}
& VIVI-Rot(4)             &       87.1 &      74.2 & 62.4 &       73.5 & 68.6 & 97.0 &   61.1 &     86.8 &    98.3 &     92.8 &        76.9 &        99.8 &       92.1 &  76.3 &       79.1 &    100.0 &        96.5 &      100.0 &       97.7 & 85.3 \\
\rotyt \cellcolor{white}
& VIVI-Rot(2)             &       86.7 &      74.1 & 61.6 &       75.1 & 67.6 & 97.0 &   61.9 &     86.7 &    98.4 &     92.6 &        77.7 &        99.8 &       92.5 &  76.4 &       81.3 &    100.0 &        96.6 &       99.9 &       97.1 & 85.4 \\
\rotytaa \cellcolor{white}
& Rot-YT-F-AA             &       86.8 &      72.5 & 63.0 &       74.7 & 68.4 & 96.9 &   60.1 &     86.4 &    98.4 &     92.8 &        78.5 &        99.8 &       92.2 &  76.4 &       81.5 &    100.0 &        96.6 &       99.7 &       98.1 & 85.4 \\
\exyt \cellcolor{white}
& Ex-YT-F                 &       85.0 &      73.6 & 63.8 &       84.9 & 70.5 & 96.8 &   60.6 &     87.2 &    98.6 &     94.3 &        78.9 &        99.8 &       93.3 &  76.8 &       80.9 &    100.0 &        96.6 &       99.9 &       97.3 & 86.2 \\
& MT-SSL                  &       88.0 &      76.1 & 64.4 &       80.0 & 72.3 & 97.2 &   63.0 &     85.8 &    98.3 &     93.7 &        78.6 &        99.7 &       93.0 &  75.4 &       80.4 &    100.0 &        96.5 &      100.0 &       98.1 & 86.3 \\
& Ex-ImageNet             &       83.5 &      74.2 & 65.4 &       83.4 & 74.9 & 96.8 &   60.4 &     85.5 &    98.7 &     94.5 &        79.8 &        99.8 &       93.5 &  75.5 &       80.4 &    100.0 &        96.5 &       99.9 &       98.0 & 86.4 \\
& Rot-ImageNet            &       88.5 &      76.4 & 67.7 &       83.0 & 73.1 & 97.0 &   63.2 &     85.4 &    98.5 &     93.9 &        79.1 &        99.9 &       92.2 &  76.0 &       82.0 &    100.0 &        96.6 &      100.0 &       98.3 & 86.9 \\
\exyt \cellcolor{white}
& VIVI-Ex(2)-Ord          &       86.0 &      75.7 & 62.1 &       87.1 & 76.1 & 96.9 &   63.7 &     87.2 &    98.6 &     94.6 &        79.9 &        99.8 &       93.5 &  76.5 &       80.9 &    100.0 &        96.5 &       99.8 &       97.9 & 87.0 \\
\exyt \cellcolor{white}
& Ex-YT-S                 &       87.4 &      75.9 & 64.8 &       85.7 & 75.0 & 96.9 &   63.2 &     87.0 &    98.6 &     94.5 &        80.1 &        99.8 &       93.4 &  77.4 &       80.4 &    100.0 &        96.6 &       99.9 &       97.3 & 87.1 \\
\exyt \cellcolor{white}
& VIVI-Ex(4)              &       86.1 &      76.3 & 61.8 &       87.3 & 76.7 & 97.0 &   64.0 &     86.9 &    98.6 &     94.7 &        80.2 &        99.8 &       93.5 &  76.8 &       81.3 &    100.0 &        96.6 &       99.8 &       98.2 & 87.1 \\
\exytaa \cellcolor{white}
& Ex-YT-F-AA              &       88.1 &      75.1 & 67.7 &       86.1 & 73.5 & 96.9 &   62.2 &     86.8 &    98.8 &     94.6 &        79.0 &        99.9 &       93.5 &  76.5 &       82.9 &    100.0 &        96.6 &       99.9 &       97.9 & 87.2 \\
\exyt \cellcolor{white}
& VIVI-Ex(2)              &       86.6 &      76.1 & 63.4 &       88.2 & 74.4 & 97.0 &   64.1 &     88.4 &    98.6 &     94.7 &        79.2 &        99.8 &       93.4 &  77.1 &       80.9 &    100.0 &        96.5 &       99.9 &       97.6 & 87.2 \\
\exytaa \cellcolor{white}
& Ex-YT-S-AA              &       89.0 &      76.5 & 67.3 &       86.2 & 75.9 & 97.0 &   63.6 &     86.9 &    98.8 &     94.6 &        80.3 &        99.8 &       93.3 &  77.1 &       82.0 &    100.0 &        96.6 &       99.9 &       97.6 & 87.5 \\
\exytaa \cellcolor{white}
& VIVI-Ex(4)-AA           &       88.8 &      76.8 & 64.0 &       87.1 & 75.9 & 97.2 &   63.9 &     88.6 &    98.6 &     94.5 &        79.5 &        99.8 &       93.2 &  76.7 &       84.0 &    100.0 &        96.6 &       99.8 &       97.6 & 87.5 \\
\excoyt \cellcolor{white}
& VIVI-Ex(4)-Co(10\%)      &       89.3 &      79.1 & 67.6 &       89.1 & 83.2 & 96.9 &   66.5 &     90.1 &    98.4 &     93.0 &        79.6 &        99.5 &       92.1 &  74.8 &       83.1 &    100.0 &        96.5 &       99.8 &       93.6 & 88.0 \\
\exyt \cellcolor{white}
& VIVI-Ex(4)-Big          &       89.1 &      79.4 & 64.7 &       89.6 & 78.7 & 97.1 &   69.2 &     86.9 &    98.6 &     95.6 &        80.2 &        99.8 &       93.6 &  77.2 &       81.8 &    100.0 &        96.6 &       99.9 &       98.6 & 88.3 \\
\exytaa \cellcolor{white}
& VIVI-Ex(4)-Big-AA       &       90.5 &      80.4 & 68.5 &       87.5 & 78.3 & 97.3 &   68.7 &     88.7 &    98.7 &     95.3 &        80.5 &        99.9 &       92.8 &  77.8 &       81.0 &    100.0 &        96.7 &      100.0 &       98.1 & 88.5 \\
& Semi-Ex-10\%             &       85.3 &      82.7 & 70.5 &       92.2 & 89.0 & 97.0 &   67.4 &     86.0 &    98.6 &     94.7 &        78.8 &        99.8 &       93.1 &  76.8 &       81.5 &    100.0 &        96.5 &      100.0 &       97.8 & 88.8 \\
\excoyt \cellcolor{white}
& VIVI-Ex(4)-Co(100\%)     &       92.5 &      82.0 & 73.2 &       92.7 & 90.9 & 96.8 &   70.7 &     87.4 &    98.5 &     93.7 &        80.2 &        99.4 &       91.2 &  73.4 &       82.1 &    100.0 &        96.5 &       98.9 &       96.5 & 89.3 \\
& Sup-100\%                &       94.1 &      83.8 & 74.0 &       93.2 & 91.9 & 97.0 &   70.7 &     83.9 &    98.8 &     95.3 &        79.3 &        99.8 &       92.1 &  76.4 &       80.7 &    100.0 &        96.4 &       99.8 &       97.7 & 89.7 \\
& Sup-Rot-100\%            &       94.6 &      84.8 & 75.9 &       94.7 & 91.5 & 97.0 &   70.2 &     85.9 &    98.8 &     94.9 &        79.5 &        99.8 &       92.5 &  76.5 &       82.3 &    100.0 &        96.5 &      100.0 &       98.4 & 90.2 \\

\excoytaa \cellcolor{white} \multirow{-27}{*}{\rotatebox{90}{full}}& VIVI-Ex(4)-Co(100\%)-Big &       93.5 &      85.9 & 77.2 &       94.4 & 91.6 & 97.3 &   73.7 &     89.4 &    98.8 &     95.1 &        81.0 &        99.7 &       92.5 &  76.7 &   84.8 &    100.0 &        96.6 &       99.7 &       94.6 & 90.7 \\
\bottomrule
\end{tabularx}

    \caption{Testing accuracy for every data set in the VTAB benchmark using 1000 and all samples for fine-tuning. Each number is the median of three fine-tuning runs. The proposed methods have the prefix \gls{vivi}. ``Ex'' and ``Rot'' stand for exemplar \cite{dosovitskiy2014discriminative} and rotation prediction \cite{gidaris2018unsupervised} frame-level self-supervision, respectively. These identifiers are followed with the number of shots in parentheses if an InfoNCE prediction loss across shots is used (except methods using shot order prediction have the suffix ``-Ord''). Baseline methods only using frames and shots have the suffix ``YT-F'' and ``YT-S'', respectively. The suffix ``-AA'' denotes methods that use \acrlong{aa} \cite{cubuk2018autoaugment}.}
    \label{tab:full-results}
\end{table*}

\FloatBarrier

\section{Architectures}
Here we expand on the short description in Section~\ref{sec:exp-setup}. The frame encoder $f$ is modelled using the ResNet-50 v2 \cite{he2016identity} architecture with BatchNorm \cite{ioffe2015batch}.
We also investigate in several experiments the effect of model capacity by widening the network by a factor of three.
To avoid mismatch in batch statistics between the two data sources, in the co-training experiments we replace the BatchNorm with GroupNorm \cite{wu2018group} and also standardize \cite{qiao2019weight} the weights of the convolutions.

For each prediction task, we attach a different linear head to the 2048-dimensional pre-logits ResNet representation before applying the respective loss or prediction function. For exemplar, following \cite{kolesnikov2019revisiting}, we use a linear head with 1000 outputs with L2-normalization of the features before feeding into the triplet-loss. For rotation prediction we rely on a linear head with 4 outputs. For the video-level loss (prediction across shots using $\Lnce$ and temporal order prediction) we project the pre-logits, average-pooled across the frames of the same shot, to 512 dimensions using a linear head, and feed this representation to the prediction functions $\gm$. Finally, in the experiments with co-training, we rely on an additional linear classification head with 1000 outputs.

For the $\Lnce$ loss, when we sample 2 shots, we predict one from the other using an \gls{mlp}, i.e., the function $g$ in \eqref{eq:infonce} has the form $g(e,e')=\phi_1(e)^\top \phi_2(e')$, where $\phi_1, \phi_2$ are \glspl{mlp} with a single hidden layer with 256 units and 128 outputs.
In the experiments with 4 shots, we use a 2-layer \gls{lstm} prediction function with 256 hidden units to predict every shot embedding from the previous ones. To match the dimension of the \gls{lstm} output (256) and that of the future shot embeddings (512) we employ another linear layer. 
We use temporal order prediction only together with exemplar-based \gls{ssl} and for data with 2 shots per video, relying on a single-hidden-layer \gls{mlp} with 512 hidden units as prediction function.

For both frame and shot-level \gls{ssl} approaches we use the augmentation mechanism from \cite{szegedy2015going}. For models co-trained with a supervised loss based on a fraction of ImageNet we additionally use the same HSV-space color randomization as \cite{zhai2019s4l}.
We also perform experiments where we replace the augmentation mechanism from \cite{szegedy2015going} with \gls{aa}, which is an augmentation policy learned using a reinforcement learning algorithm from the full ImageNet data set. More specifically, we rely on the TF-Hub module publicly available at \url{https://tfhub.dev/google/image_augmentation/nas_imagenet/1}.

\section{Training details}

Table~\ref{tab:train-details} provides details about the schedules, batch size, loss weights, etc. used for the individual methods. When exploring the effect of \gls{aa} we reduce the weight of the video-level loss, $\lambda$, by a factor of 2. The schedule for VIVI-Ex(4)-Co(10\%) is motivated as follows. We take the schedule and batch size used for the ImageNet exemplar co-training experiments for 10\% labeled ImageNet examples from \cite{zhai2019s4l}, stretch the schedule to 100k iterations and reduce the batch size (as well as the learning rate) so that number of epochs over the 10\% (128k example) data set matches that of \cite{zhai2019s4l}. Table~\ref{tab:shot-stat} shows some statistics of the \gls{yt8m} subset we use for training.

We set the margin parameter in the semi-hard triplet loss \cite{schroff2015facenet} to 0.5. For rotation-based \gls{ssl}, following common practice \cite{gidaris2018unsupervised, kolesnikov2019revisiting}, we compute the predicted rotation after appending to the mini-batch 3 rotated copies of the mini-batch along the batch dimension and compute the rotation loss for all rotated copies.

We train all models on 128 cores of a Google TPU v3 Pod. For exemplar \gls{ssl} the triplet loss is computed per core. For all frame/shot level loss variants, $\Lnce$ is computed across all cores when prediction is across 4 shots, and computed per core when prediction is across 2 shots as computing the loss across all cores led to instabilities for that case.

\begin{table}[h]
    \centering
    \centering
    \small
\newcommand{\green}[1]{\textcolor{Green}{(#1)}}
\begin{tabular}{lll}
\toprule
& \textsc{Mean} & \textsc{Std.} \\
\midrule
Number of shots per video & \; 25.5 & 30.3 \\
Shot duration & \quad 9.0 sec. & 25.3 sec. \\
Video duration & 230.7 sec. & 61.7 sec. \\
\bottomrule
\end{tabular}
    \caption{Statistics of the \gls{yt8m} subset we use for training.}
    \label{tab:shot-stat}
\end{table}

\begin{table}
    \small
    \makebox[\textwidth][c]{
    \centering
    \small
\setlength{\tabcolsep}{3pt}
\setlength{\extrarowheight}{3.5pt}
\renewcommand{\arraystretch}{0.75}
\begin{tabular}{lccccccc P{1.5cm} c}
\toprule
& LR & \#it. & w. \#it. & LR schedule & WD & $\lambda$ & $\gamma$ & batch size & \#exemp.\\
\midrule
Ex-ImageNet & 0.8 & 120k & 17k & $\times$0.1@52k;86k & $10^{-4}$ & - & - & 2048 & 8 \\
Ex-YT-F & 0.8 & 120k & 17k & $\times$0.1@52k;86k & $10^{-4}$ & - & - &  2048 & 8 \\
Ex-YT-S & 0.8 & 120k & 5k & $\times$0.1@90k;110k & $10^{-4}$ & - & - &  2048 & 8 (sh.) \\
VIVI-Ex(2)-Ord & 0.8 & 120k & 5k & $\times$0.1@90k;110k & $10^{-4}$ & $\{2.0, 1.0, \underline{0.5}\}$ & - &  $1024 \cdot 2$ sh. & 8 (sh.) \\
VIVI-Ex(2) & 0.8 & 120k & 5k & $\times$0.1@90k;110k & $10^{-4}$ & $\{0.08, 0.04, \underline{0.02}\}$ & - &  $1024 \cdot 2$ sh. & 8 (sh.) \\
VIVI-Ex(4) & 0.8 & 120k & 5k & $\times$0.1@90k;110k & $10^{-4}$ & $\{\underline{0.04}, 0.02, 0.01\}$ & - &  $512 \cdot 4$ sh. & 8 (sh.) \\
VIV-Ex(4)-Big & 0.8 & 120k & 5k & $\times$0.1@90k;110k & $10^{-4}$ & 0.04 & - &  $512 \cdot 4$ sh. & 8 (sh.) \\
\midrule
VIVI-Ex(4)-Co(10\%) & 0.1 & 100k & 3k & $\times$0.1@76k;88k;96k & $10^{-3}$ & 0.04 & $\{1.0, \underline{4.0}, 8.0, 16.0\}$ & $512 \cdot 4$ sh., 256 im. & 8 (sh.) \\
VIVI-Ex(4)-Co(100\%) & 0.8 & 100k & 3k & $\times$0.1@70k;85k;95k & $10^{-4}$ & 0.04 & $\{0.1, 0.5, \underline{1.0}, 5.0\}$ & $512 \cdot 4$ sh., 2048 im. & 8 (sh.) \\
VIVI-Ex(4)-Co(100\%)-Big & 0.8 & 100k & 3k & $\times$0.1@70k;85k;95k & $10^{-4}$ & 0.04 & $\{0.1, 0.5, 1.0, \underline{5.0}\}$ & $512 \cdot 4$ sh., 2048 im. & 8 (sh.) \\
\midrule
Rot-ImageNet & 0.8 & 120k & 17k & $\times$0.1@52k;86k & $10^{-4}$ & - & - & 2048 & 1 \\
Rot-YT-F & 0.8 & 120k & 17k & $\times$0.1@52k;86k & $10^{-4}$ & - & - & 2048 & 1 \\
VIVI-Rot(2) & 0.8 & 120k & 5k & $\times$0.1@90k;110k & $10^{-4}$ & $\{0.2, 0.1, \underline{0.05}, 0.025\}$ & - & $1024 \cdot 2$ sh. & 4 (sh.) \\
VIVI-Rot(4) & 0.8 & 120k & 5k & $\times$0.1@90k;110k & $10^{-4}$ & $\{0.32, 0.16, \underline{0.08}, 0.04\}$ & - & $512 \cdot 4$ sh. & 4 (sh.) \\
\bottomrule
\end{tabular}
    }
    \caption{Learning rate (LR), number of training iterations (\#it.), number of linear warm-up iterations (w. \#it.), learning rate schedule (LR schedule), weight decay (WD), video-level loss weight ($\lambda$), supervised cross-entropy loss weight ($\gamma$), batch size, and the number of exemplars (\#exemp.) for the different models considered in this paper. Lists of values indicate values explored in the parameter sweep, with the optimal value (in terms of validation VTAB 1000 example score) underlined. For the co-training methods we indicate video (suffix ``sh.'') and image (suffix ``im.'') batch size. If the number of exemplars is followed by ``(sh.)'' we use consecutive frames of the same shot to create exemplars.}
    \label{tab:train-details}
\end{table}

\section{Baseline fine-tuning details}

As mentioned in the main manuscript we compared against two baseline methods: MT-SSL (Multi-Task Self-Supervised Learning)~\cite{doersch2017multi}, and TI (Transitive Invariance)~\cite{wang2017transitive}. For MT-SSL we considered two variants: MS which was pre-trained on motion segmentation only, and MT-SSL which combined MS with three other tasks in a multi-task setting. We obtained pre-trained checkpoints for all three methods (MS, MT-SSL, and TI) from the authors of their respective prior works.

\subsection{Fine-tuning motion segmentation and multi-task SSL baselines}
MS and MT-SSL pre-trained a ResNet-101 up to block3. The representation at block3 is $7 \times 7 \times 1024$, which is too big. In~\cite{doersch2017multi}, the authors used max-pooling to down-sample this to $3 \times 3 \times 1024$ and then trained a linear predictor for ImageNet classification. We experimented with this approach for VTAB evaluation. The default evaluation protocol for VTAB is to sweep over initial learning rates: $0.1$ and $0.01$. These were too high for the MS and MT-SSL models. For several downstream evaluation tasks fine-tuning diverged. We therefore modified the evaluation sweep minimally to sweep over initial learning rates: $0.1, 0.05, 0.01$. We also evaluated a simpler alternative: Global average pooling the block3 representation into a $1 \times 1 \times 1024$ dimensional vector. We found that global average pooling the representation achieved best results on the VTAB validation set. It also did not diverge at higher learning rates,  so we could use the default learning rate schedule in this case. We therefore used this setting for the final evaluation on test data.

\subsection{Fine-tuning the transitive invariance baseline}
We exported the pre-trained caffe checkpoint into TensorFlow using the Caffe-TensorFlow tool\footnote{\url{https://github.com/ethereon/caffe-tensorflow}}. We found that the pre-trained VGG-16 backbone diverges at higher learning rates when fine-tuning downstream on VTAB tasks. We therefore manually adjusted the sweep over initial learning rates and found $0.01, 0.005, 0.001$ to work well. Another challenge with transferring this baseline model to several downstream data sets was that it is a patch-based model that expects $96 \times 96$ dimensional input, whereas the VTAB benchmark scales all images to $224 \times 224$. We experimented with three ways of deploying this downstream: (a) Resize the input image from $224 \times 224$ into $96 \times 96$, (b) apply the model fully convolutionally and compute a global average pool at the end, and (c) crop patches of size $96 \times 96$ at stride $32$ from the input image and then average the representations across all of these. We found that (c) was computationally extremely expensive. (b) performed best and we report results for that approach on the VTAB test set.

\clearpage

\FloatBarrier

\section{Additional results}

\paragraph{Object detection} We evaluate the proposed framework as a pre-training step for object detection. Specifically, we use different VIVI models and baselines as backbone for RetinaNet \cite{lin2017focal} and evaluate it on the COCO-2017 data set \cite{lin2014microsoft}. For all experiments, we rely on the standard ResNet-50 v2 architecture. We use the standard RetinaNet training protocol with fine-tuning and only adapt the learning rate in cases where the default learning rate is too high by halving it until the training loss decays smoothly (i.e., we tune the learning rate based on the training loss). Table~\ref{tab:coco-detection} shows the standard COCO-2017 detection metric for different models. It can be seen that VIVI-Ex(4) outperforms the Ex-YT-F, MS, and MT-SSL baselines, and VIVI-Ex(4)-Co(100\%) improves over supervised ImageNet pre-training (Sup-100\%).

\begin{table}[h!]
    \centering
    \small
\begin{tabular}{ll}
\toprule
\textsc{Method} & \textsc{AP} \\
\midrule
RetinaNet (Random init.) & 26.9 \\
RetinaNet (MS) & 32.3 \\
RetinaNet (Ex-YT-F) & 32.6 \\
RetinaNet (MT-SSL) & 32.7 \\
RetinaNet (VIVI-Ex(4)) & 33.6 \\
RetinaNet (Sup-100\%) & 35.3 \\
RetinaNet (VIVI-Ex(4)-Co(100\%)) & 36.5 \\
\bottomrule
\end{tabular}
    \caption{Object detection performance of RetinaNet \cite{lin2017focal} on the COCO-2017 data set \cite{lin2014microsoft} for different ways of pre-training the ResNet-50 v2 backbone (Sup-100\% corresponds to standard supervised ImageNet pre-training). The reported AP values are the median over 3 runs.}
    \label{tab:coco-detection}
\end{table}

\begin{figure}[h]
\centering
\includegraphics[width=0.75\textwidth]{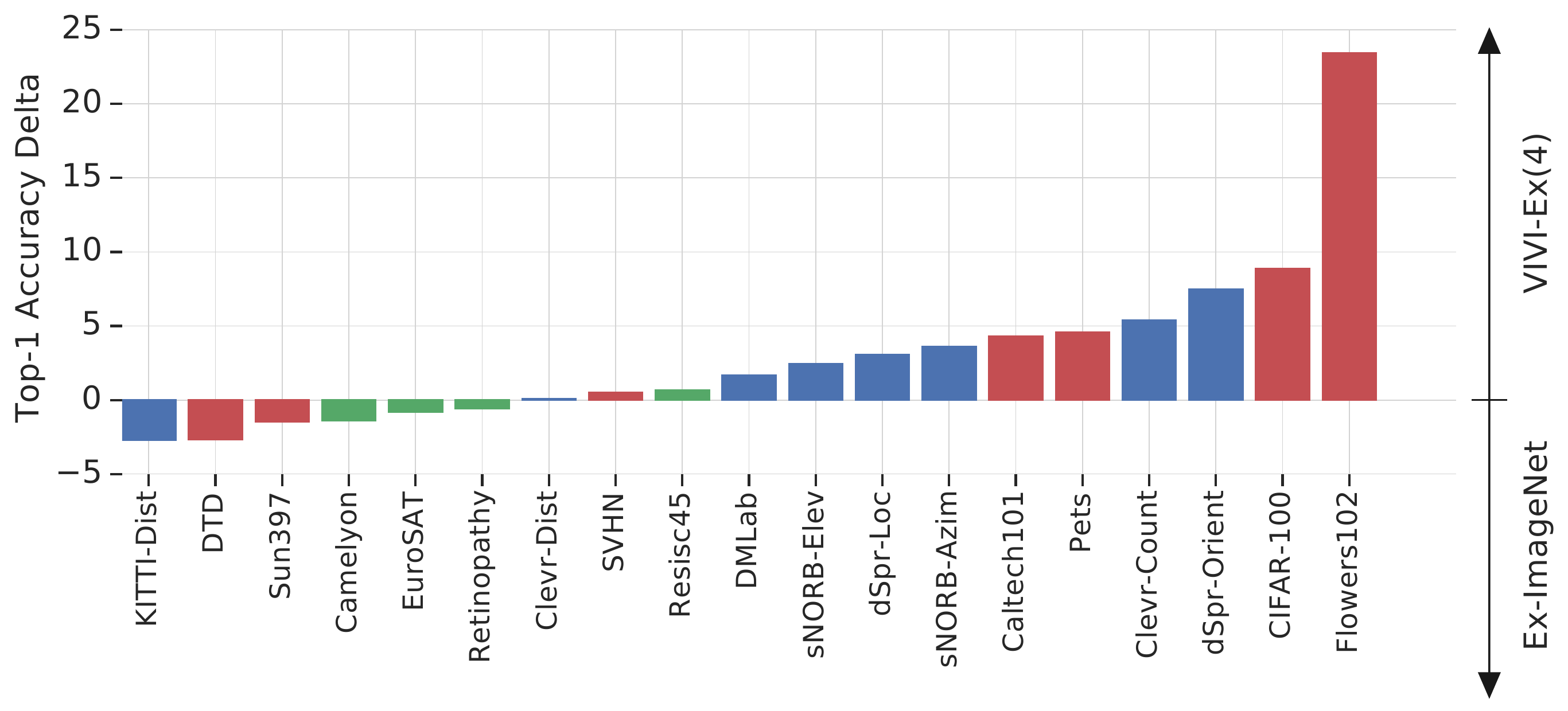}
\caption{Per-data set comparison of ImageNet-based exemplar \gls{ssl} (Ex-ImageNet) with VIVI-Ex(4). Training on \gls{yt8m} rather than ImageNet and exploiting temporal information mostly helps on natural (red) and structured (blue) data sets, and slightly hurts for some specialized (green) data sets.}
\label{fig:atari2}
\end{figure}

\begin{figure}[h]
\centering
\includegraphics[width=0.75\textwidth]{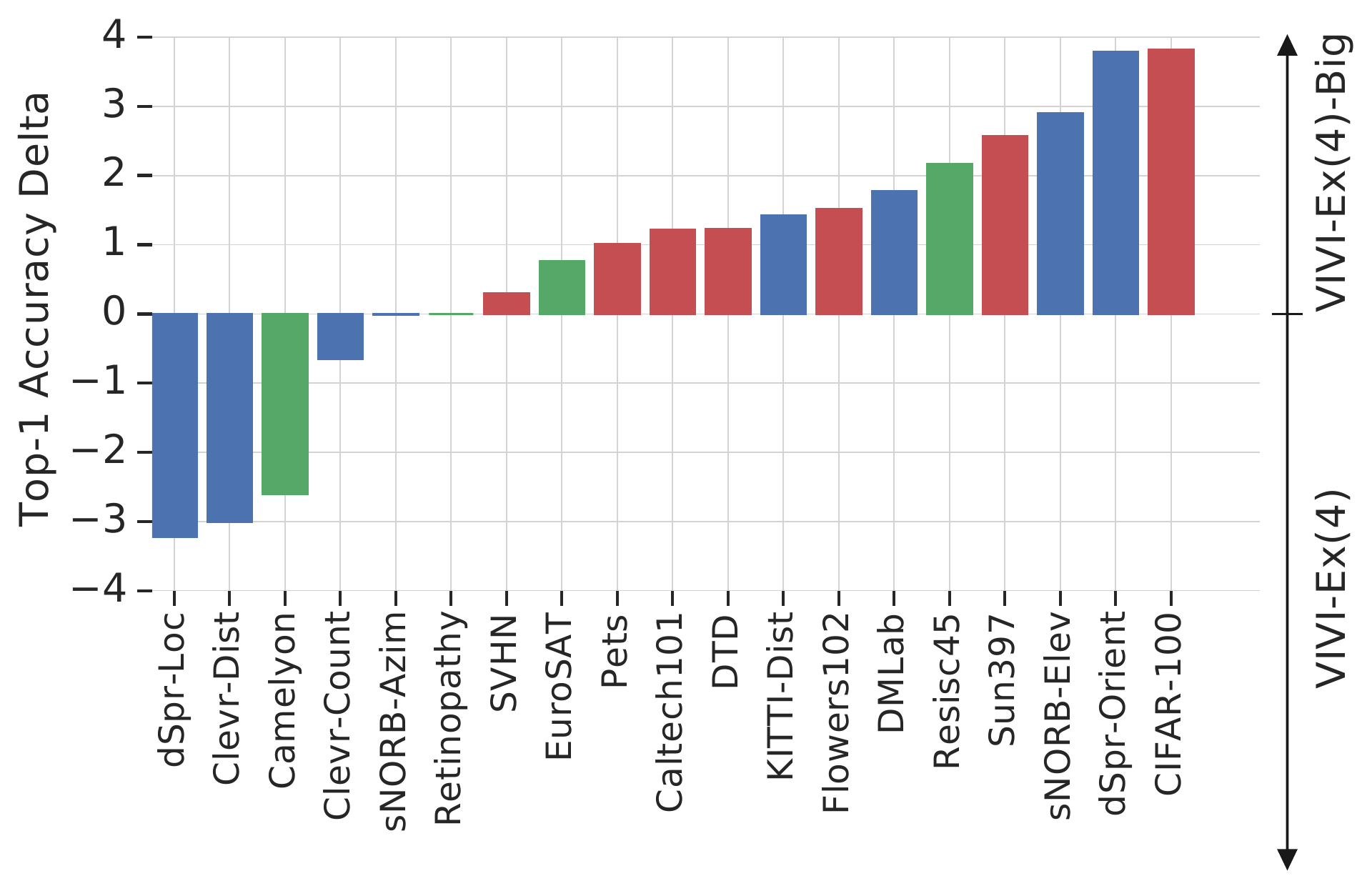}
\caption{Per-data set comparison of VIVI-Ex(4) and a 3$\times$ wider counterpart (VIVI-Ex(4)-Big). Increasing model capacity leads to an increase in accuracy for all natural (red) data sets and some structured (blue) and specialized (green) data sets. However, some structured and specialized data sets also incur a reduction in accuracy.}
\label{fig:atari3}
\end{figure}

\begin{figure}[h]
\centering
\includegraphics[width=0.75\textwidth]{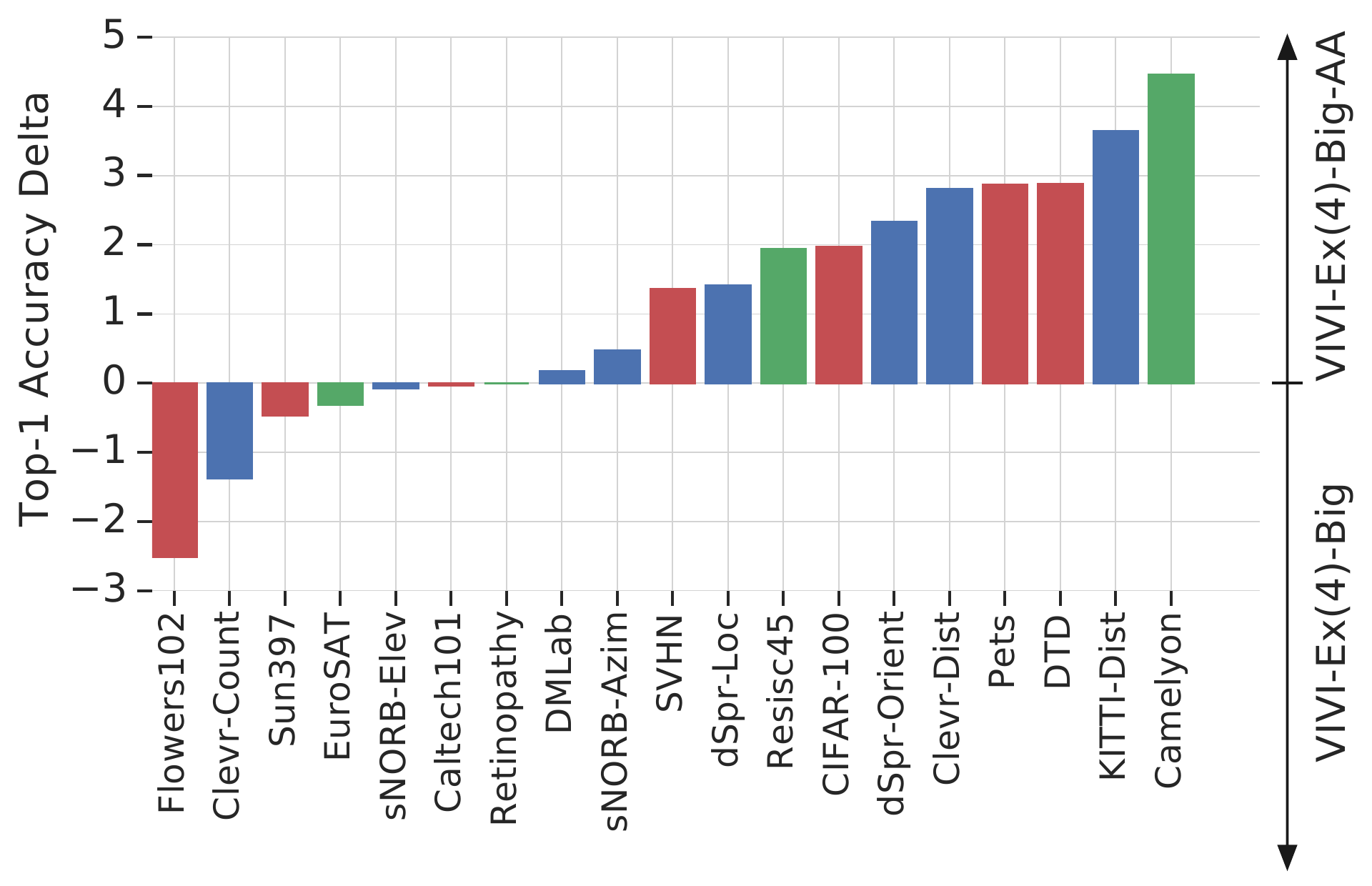}
\caption{Per-data set comparison of VIVI-Ex(4) and a variant using \gls{aa}. \gls{aa} seems to benefit all data set categories similarly, and also leads to reductions in accuracy for a few data sets from all categories.}
\label{fig:atari4}
\end{figure}

\begin{figure}[h]
\centering
\includegraphics[width=0.75\textwidth]{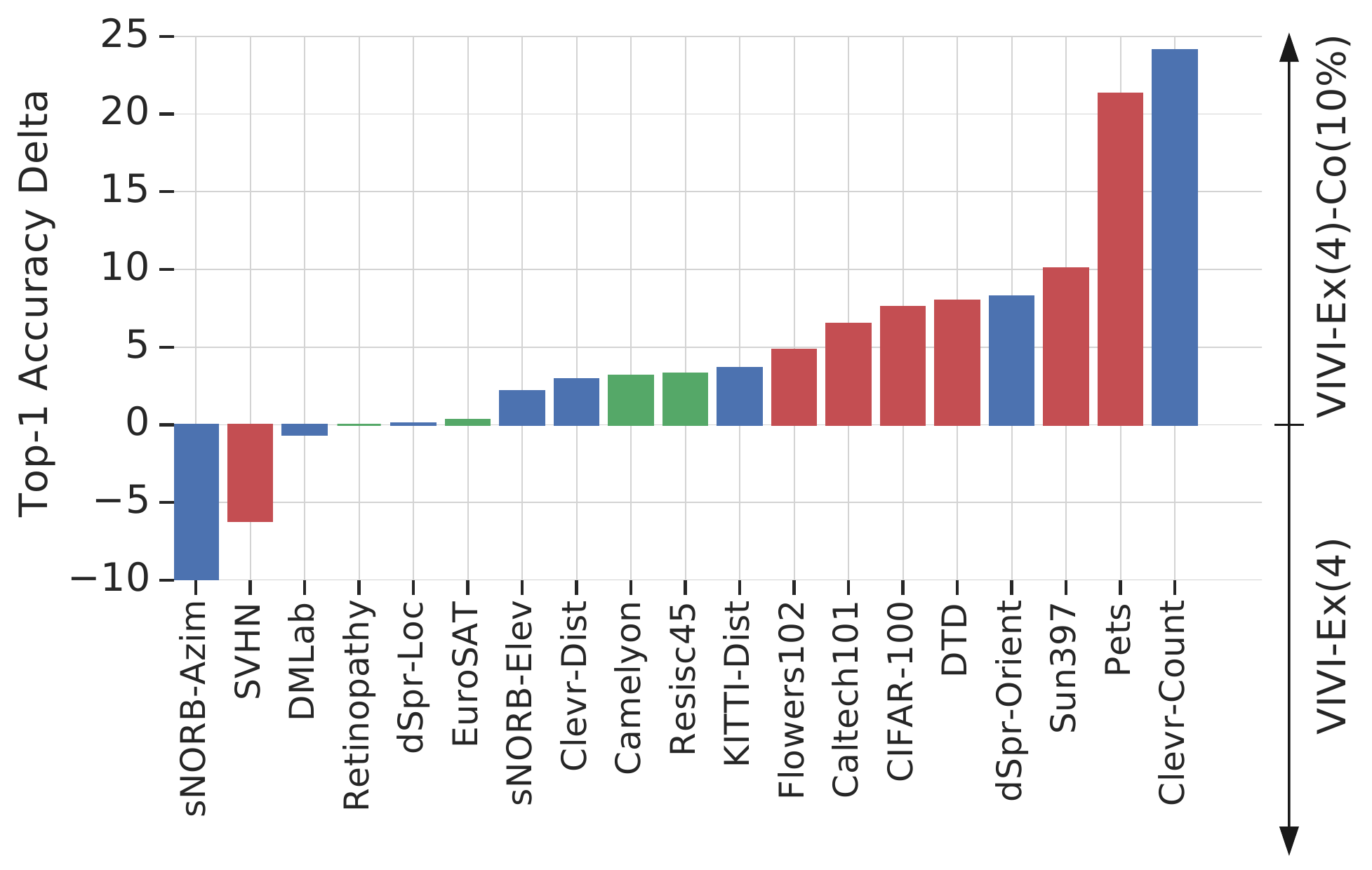}
\caption{Per-data set comparison of VIVI-Ex(4) and its counterpart co-trained with 10\% class-balanced ImageNet data (VIVI-Ex(4)-Co(10\%)). Most data sets from each category incur an increase in accuracy, but one data set from each the natural and structured categories suffer a significant loss in accuracy.}
\label{fig:atari5}
\end{figure}

\begin{figure}[h]
\centering
\includegraphics[width=0.75\textwidth]{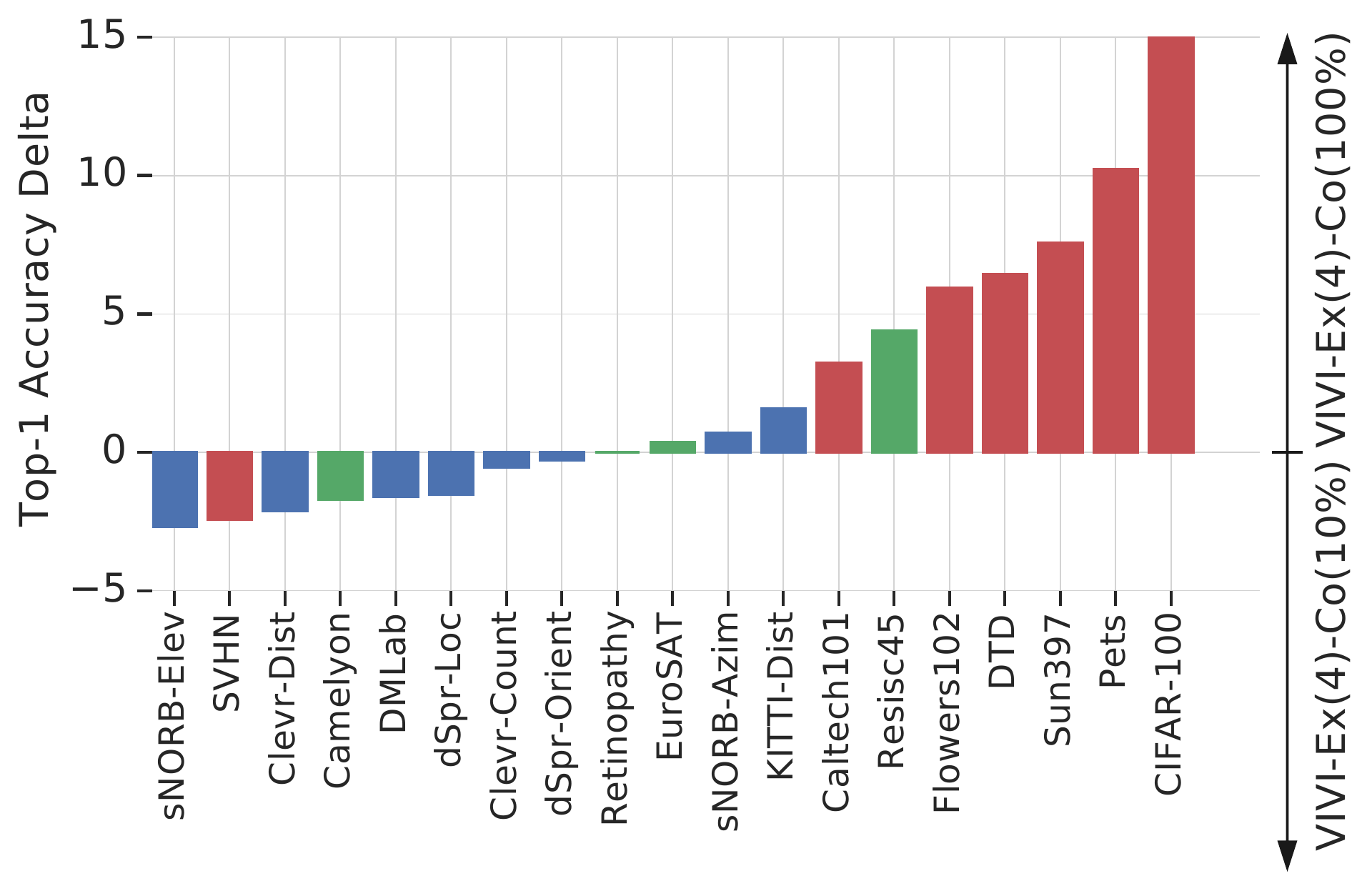}
\caption{Effect of increasing the number of ImageNet images used for co-training from 10\% (VIVI-Ex(4)-Co(10\%)) to 100\% (VIVI-Ex(4)-Co(100\%)). The accuracy on the majority of natural (red) data sets is significantly increased, whereas most of the structured data sets incur a slight drop in accuracy.}
\label{fig:atari6}
\end{figure}

\paragraph{Additional figures} In Fig.~\ref{fig:atari2} to \ref{fig:atari6} we provide per-data set comparisons of different model pairs to better understand the effect of increasing the model size, using \gls{aa}, and co-training with different amounts of labeled images. All numbers are accuracies when using 1000 labels for fine-tuning.

\FloatBarrier

\FloatBarrier

\section{Evaluation on VTAB Version 2} \label{sec:vtab-v2}

In Table~\ref{tab:full-results-v2} we report complete testing results for the 1000 example transfer regime of Version 2 (arXiv:1910.04867v2) of the VTAB benchmark \cite{zhai2019visual}. Table~\ref{tab:per-group-results-v2} shows the mean testing accuracy per data set category, and Table~\ref{tab:cotraining-v2} is the Version 2-analog of Table~\ref{tab:cotraining}. Note that the full data set evaluation is the same in both Versions 1 and 2. The 1000 example regime of Version 2 uses 800 samples for fine-tuning and 200 samples for validation to select the hyper-parameters for fine-tuning (see Section~\ref{sec:exp-setup}; the considered set of fine-tuning hyper-parameters remains the same). In contrast, Version 1 uses all the 1000 samples for fine-tuning, and the standard validation set for each data set. We emphasize that we do not re-select the \emph{training} hyper-parameters (such as the video-level loss weight, see Table~\ref{tab:train-details}), but only the hyper-parameters for the transfer step.

It can be seen that the relative improvements of our methods in terms of the VTAB 1000 example mean accuracy over the baselines obtained  for Version 2 are comparable to Version 1, with few exceptions. Accordingly, the ranking of methods according to the mean accuracy remains mostly unchanged.

\begin{table*}[h!]
    \centering
    \small
    \newcommand{\rotyt}{\rowcolor{orange!20!white}}
\newcommand{\rotytaa}{\rowcolor{orange!40!white}}
\newcommand{\exyt}{\rowcolor{violet!20!white}}
\newcommand{\exytaa}{\rowcolor{violet!40!white}}
\newcommand{\excoyt}{\rowcolor{yellow!20!white}}
\newcommand{\excoytaa}{\rowcolor{yellow!40!white}}
\fontsize{7pt}{7pt}\selectfont
\newcolumntype{C}{>{\centering\arraybackslash}X}
\setlength{\tabcolsep}{0pt}
\setlength{\extrarowheight}{5pt}
\renewcommand{\arraystretch}{0.80}
\begin{tabularx}{\linewidth}{p{10pt}p{2.8cm}!{\color{lightgray}\vline} CCCCCCC!{\color{lightgray}\vline}CCCC!{\color{lightgray}\vline}CCCCCCCC!{\color{lightgray}\vline}C}
\toprule
 &
 & \rotatebox{90}{\tikz\fill[natural] (0,0) circle (.5ex); Caltech101}
 & \rotatebox{90}{\tikz\fill[natural] (0,0) circle (.5ex); CIFAR-100}
 & \rotatebox{90}{\tikz\fill[natural] (0,0) circle (.5ex); DTD}
 & \rotatebox{90}{\tikz\fill[natural] (0,0) circle (.5ex); Flowers102}
 & \rotatebox{90}{\tikz\fill[natural] (0,0) circle (.5ex); Pets}
 & \rotatebox{90}{\tikz\fill[natural] (0,0) circle (.5ex); SVHN}
 & \rotatebox{90}{\tikz\fill[natural] (0,0) circle (.5ex); Sun397}
 & \rotatebox{90}{\tikz\fill[specialized] (0,0) circle (.5ex); Camelyon}
 & \rotatebox{90}{\tikz\fill[specialized] (0,0) circle (.5ex); EuroSAT}
 & \rotatebox{90}{\tikz\fill[specialized] (0,0) circle (.5ex); Resisc45}
 & \rotatebox{90}{\tikz\fill[specialized] (0,0) circle (.5ex); Retinopathy}
 & \rotatebox{90}{\tikz\fill[structured] (0,0) circle (.5ex); Clevr-Count}
 & \rotatebox{90}{\tikz\fill[structured] (0,0) circle (.5ex); Clevr-Dist}
 & \rotatebox{90}{\tikz\fill[structured] (0,0) circle (.5ex); DM-Lab}
 & \rotatebox{90}{\tikz\fill[structured] (0,0) circle (.5ex); KITTI-Dist} 
 & \rotatebox{90}{\tikz\fill[structured] (0,0) circle (.5ex); dSpr-Loc}
 & \rotatebox{90}{\tikz\fill[structured] (0,0) circle (.5ex); dSpr-Ori}
 & \rotatebox{90}{\tikz\fill[structured] (0,0) circle (.5ex); sNORB-Azim}
 & \rotatebox{90}{\tikz\fill[structured] (0,0) circle (.5ex); sNORB-Elev}
 & \rotatebox{90}{\tikz\fill[all] (0,0) circle (.5ex); Mean} \\
 
\midrule

& TI                      &       54.9 &       7.1 & 38.3 &       28.2 & 32.3 & 77.0 &    7.4 &     50.0 &    84.1 &     50.0 &        63.1 &        12.7 &       61.7 &  35.0 &       41.6 &     86.1 &        59.3 &       21.1 &       29.2 & 44.2 \\
& MS                      &       52.3 &      12.7 & 37.3 &       32.6 & 15.8 & 81.8 &    6.8 &     76.8 &    89.7 &     49.7 &        57.3 &        43.2 &       55.7 &  38.4 &       48.4 &     81.2 &        46.4 &       34.8 &       35.1 & 47.1 \\
& Rel.Pat.Loc             &       67.8 &      17.9 & 51.0 &       67.2 & 38.8 & 61.7 &   10.5 &     73.4 &    92.6 &     66.2 &        59.7 &        44.3 &       55.7 &  39.4 &       57.8 &     63.6 &        32.7 &       29.9 &       34.9 & 50.8 \\
& Jigsaw                  &       66.2 &      14.8 & 50.7 &       65.3 & 34.0 & 54.9 &   11.4 &     73.0 &    91.5 &     66.7 &        71.3 &        44.1 &       56.2 &  42.2 &       63.8 &     66.0 &        34.2 &       32.9 &       31.7 & 51.1 \\
\rotyt \cellcolor{white} & Rot-YT-F                &       69.5 &      21.5 & 47.2 &       44.0 & 34.2 & 88.4 &    9.5 &     82.5 &    92.1 &     59.4 &        70.5 &        47.1 &       57.4 &  46.2 &       71.5 &     92.0 &        49.9 &       36.8 &       37.1 & 55.6 \\
\rotyt \cellcolor{white} & VIVI-Rot(4)             &       72.7 &      24.6 & 48.6 &       48.9 & 39.1 & 88.8 &   11.0 &     83.1 &    93.5 &     65.9 &        71.2 &        49.0 &       58.6 &  46.5 &       70.4 &     89.1 &        51.7 &       24.7 &       39.8 & 56.7 \\
\rotyt \cellcolor{white} & VIVI-Rot(2)             &       73.0 &      26.7 & 49.1 &       55.0 & 43.2 & 88.4 &   12.7 &     83.0 &    93.1 &     66.7 &        71.6 &        45.1 &       57.8 &  45.5 &       69.5 &     88.0 &        51.1 &       37.5 &       38.1 & 57.6 \\
\rotytaa \cellcolor{white} & Rot-YT-F-AA             &       77.1 &      23.0 & 50.8 &       55.3 & 35.2 & 88.6 &    9.4 &     81.9 &    93.4 &     64.5 &        71.4 &        42.8 &       58.6 &  46.9 &       71.7 &     91.4 &        53.9 &       40.7 &       39.8 & 57.7 \\
\exyt \cellcolor{white} & Ex-YT-F                 &       72.2 &      23.9 & 51.9 &       50.1 & 44.7 & 88.1 &   15.3 &     82.5 &    95.3 &     69.7 &        70.6 &        47.6 &       59.5 &  45.7 &       72.3 &     90.3 &        46.8 &       32.6 &       44.7 & 58.1 \\
& BigBiGAN                &       80.8 &      39.2 & 56.6 &       77.9 & 44.4 & 76.8 &   20.3 &     77.4 &    95.6 &     74.0 &        69.3 &        53.9 &       55.6 &  38.7 &       71.4 &     70.6 &        46.7 &       27.2 &       46.3 & 59.1 \\
& Ex-ImageNet             &       72.7 &      23.2 & 54.5 &       68.7 & 48.7 & 88.4 &   14.4 &     80.3 &    95.5 &     73.8 &        74.1 &        44.9 &       60.2 &  45.7 &       69.6 &     89.3 &        45.8 &       34.2 &       40.7 & 59.2 \\
& MT-SSL                  &       76.2 &      26.2 & 49.3 &       63.5 & 48.5 & 89.1 &   10.6 &     80.3 &    93.3 &     70.2 &        71.7 &        55.6 &       62.1 &  44.3 &       76.3 &     86.6 &        43.2 &       39.1 &       38.9 & 59.2 \\
& Rot-ImageNet            &       77.1 &      25.2 & 55.9 &       71.2 & 44.2 & 88.5 &   14.4 &     81.1 &    93.5 &     70.1 &        64.1 &        48.0 &       59.6 &  46.9 &       74.5 &     92.7 &        49.9 &       37.4 &       38.9 & 59.6 \\
\exyt \cellcolor{white} & Ex-YT-S                 &       75.4 &      28.1 & 50.2 &       74.3 & 52.3 & 88.2 &   14.7 &     82.2 &    94.5 &     74.4 &        74.0 &        52.7 &       59.9 &  45.4 &       70.7 &     88.0 &        48.3 &       32.9 &       36.6 & 60.2 \\
\exyt \cellcolor{white} & VIVI-Ex(2)-TimeArrow    &       74.9 &      28.5 & 48.3 &       77.1 & 55.3 & 88.0 &   14.1 &     83.0 &    94.1 &     73.9 &        71.5 &        46.8 &       58.2 &  46.4 &       73.9 &     92.7 &        48.1 &       35.6 &       42.3 & 60.7 \\
\exyt \cellcolor{white} & VIVI-Ex(4)              &       75.0 &      29.3 & 49.0 &       77.8 & 55.3 & 87.9 &   15.1 &     80.3 &    94.2 &     73.9 &        70.4 &        55.2 &       58.8 &  46.4 &       70.9 &     91.3 &        51.4 &       34.9 &       42.1 & 61.0 \\
\exyt \cellcolor{white} & VIVI-Ex(2)              &       74.8 &      28.8 & 49.5 &       77.5 & 54.0 & 88.2 &   14.0 &     80.0 &    94.0 &     73.7 &        71.9 &        56.0 &       61.1 &  46.5 &       70.1 &     91.8 &        51.0 &       38.2 &       45.0 & 61.4 \\
\exytaa \cellcolor{white} & Ex-YT-S-AA              &       78.0 &      29.4 & 50.7 &       75.3 & 55.3 & 89.7 &   14.0 &     83.9 &    94.9 &     76.3 &        71.7 &        52.3 &       58.7 &  46.5 &       72.2 &     91.0 &        52.6 &       35.3 &       39.9 & 61.5 \\
& Sup-10\%                 &       84.9 &      46.1 & 55.4 &       82.8 & 80.2 & 86.9 &   24.9 &     80.6 &    95.2 &     73.1 &        71.9 &        39.4 &       55.7 &  43.5 &       63.3 &     76.0 &        45.9 &       30.9 &       32.9 & 61.6 \\
\exytaa \cellcolor{white} & VIVI-Ex(4)-AA           &       78.6 &      30.5 & 50.9 &       74.8 & 55.8 & 88.8 &   14.5 &     81.2 &    94.9 &     75.5 &        71.8 &        54.3 &       61.0 &  45.7 &       76.3 &     93.0 &        54.1 &       31.7 &       39.7 & 61.7 \\
\exyt \cellcolor{white} & VIVI-Ex(4)-Big          &       76.5 &      33.0 & 51.6 &       75.2 & 57.1 & 87.5 &   17.1 &     80.1 &    94.7 &     75.1 &        72.4 &        55.4 &       57.1 &  46.6 &       73.3 &     89.4 &        55.1 &       35.6 &       44.1 & 61.9 \\
\exytaa \cellcolor{white} & Ex-YT-F-AA              &       76.5 &      27.8 & 54.7 &       75.0 & 52.0 & 89.7 &   16.7 &     83.6 &    95.5 &     74.1 &        73.5 &        51.1 &       57.9 &  48.6 &       74.8 &     94.2 &        52.7 &       34.8 &       43.9 & 61.9 \\
\exytaa \cellcolor{white} & VIVI-Ex(4)-Big-AA       &       79.7 &      35.1 & 56.0 &       72.3 & 56.9 & 89.5 &   18.0 &     82.0 &    95.2 &     77.2 &        71.6 &        54.7 &       60.3 &  48.7 &       75.8 &     92.4 &        58.5 &       35.6 &       46.2 & 63.5 \\
& Semi-Ex-10\%             &       87.7 &      52.8 & 60.4 &       84.0 & 84.0 & 87.1 &   29.2 &     79.1 &    94.9 &     77.1 &        70.1 &        39.4 &       56.0 &  42.3 &       72.0 &     73.7 &        52.3 &       37.6 &       33.5 & 63.9 \\
& Sup-100\%                &       89.8 &      54.6 & 65.6 &       88.4 & 89.1 & 86.3 &   34.5 &     79.7 &    95.3 &     81.0 &        72.6 &        41.8 &       52.5 &  42.7 &       75.3 &     81.0 &        47.3 &       32.6 &       35.8 & 65.6 \\
\excoyt \cellcolor{white} & VIVI-Ex(4)-Co(10\%)      &       82.7 &      36.5 & 57.9 &       82.0 & 76.6 & 82.2 &   24.0 &     84.7 &    94.8 &     76.8 &        73.2 &        75.5 &       60.5 &  46.7 &       77.4 &     95.0 &        58.3 &       30.5 &       45.3 & 66.3 \\
& Sup-Rot-100\%            &       89.9 &      52.8 & 68.6 &       90.3 & 88.8 & 88.7 &   32.5 &     80.5 &    95.9 &     83.4 &        73.2 &        48.2 &       57.0 &  48.5 &       79.1 &     92.5 &        50.0 &       30.6 &       32.8 & 67.5 \\
\excoyt \cellcolor{white} & VIVI-Ex(4)-Co(100\%)     &       86.7 &      51.6 & 64.8 &       88.2 & 86.3 & 80.4 &   32.5 &     84.3 &    95.1 &     80.9 &        72.6 &        78.0 &       60.4 &  45.4 &       80.2 &     92.9 &        61.7 &       30.2 &       41.8 & 69.2 \\
\excoytaa \cellcolor{white} \multirow{-29}{*}{\rotatebox{90}{1000}} & VIVI-Ex(4)-Co(100\%)-Big &       87.6 &      53.6 & 68.9 &       89.9 & 87.0 & 84.1 &   34.2 &     85.9 &    95.5 &     81.6 &        71.9 &        75.8 &       62.8 &  45.9 &   83.9 &     95.3 &        62.1 &       27.1 &       45.3 & 70.4 \\
\bottomrule
\end{tabularx}

    \caption{Testing accuracy for fine-tuning hyper-parameter selection according to {\bf Version 2 (arXiv:1910.04867v2) of the VTAB benchmark} \cite{zhai2019visual}. Each number is the median of three fine-tuning runs. The proposed methods have the prefix \gls{vivi}. ``Ex'' and ``Rot'' stand for exemplar \cite{dosovitskiy2014discriminative} and rotation prediction \cite{gidaris2018unsupervised} frame-level self-supervision, respectively. These identifiers are followed with the number of shots in parentheses if an InfoNCE prediction loss across shots is used (except methods using shot order prediction have the suffix ``-Ord''). Baseline methods only using frames and shots have the suffix ``YT-F'' and ``YT-S'', respectively. The suffix ``-AA'' denotes methods that use \acrlong{aa} \cite{cubuk2018autoaugment}.}
    \label{tab:full-results-v2}
\end{table*}

\begin{table*}[h!]
    \centering
    \small
    \begin{tabular}{lrrrr}
\toprule
\textsc{Method}  &     \textsc{Mean} & \textsc{Nat.} & \textsc{Spec.} & \textsc{Str.} \\
\midrule
TI                      &     44.2 &     35.0 &        61.8 &       43.3 \\
MS                      &     47.1 &     34.2 &        68.4 &       47.9 \\
Rel.Pat.Loc             &     50.8 &     45.0 &        73.0 &       44.8 \\
Jigsaw                  &     51.1 &     42.5 &        75.6 &       46.4 \\
Rot-YT-F                &     55.6 &     44.9 &        76.1 &       54.8 \\
VIVI-Rot(4)             &     56.7 &     47.7 &        78.4 &       53.7 \\
VIVI-Rot(2)             &     57.6 &     49.7 &        78.6 &       54.1 \\
Rot-YT-F-AA             &     57.7 &     48.5 &        77.8 &       55.7 \\
Ex-YT-F                 &     58.1 &     49.4 &        79.5 &       54.9 \\
BigBiGAN                &     59.1 &     56.6 &        79.1 &       51.3 \\
Ex-ImageNet             &     59.2 &     52.9 &        80.9 &       53.8 \\
MT-SSL                  &     59.2 &     51.9 &        78.9 &       55.8 \\
Rot-ImageNet            &     59.6 &     53.8 &        77.2 &       56.0 \\
Ex-YT-S                 &     60.2 &     54.8 &        81.3 &       54.3 \\
VIVI-Ex(2)-TimeArrow    &     60.7 &     55.2 &        80.6 &       55.5 \\
VIVI-Ex(4)              &     61.0 &     55.6 &        79.7 &       56.4 \\
VIVI-Ex(2)              &     61.4 &     55.3 &        79.9 &       57.5 \\
Ex-YT-S-AA              &     61.5 &     56.1 &        81.7 &       56.0 \\
Sup-10\%                 &     61.6 &     65.9 &        80.2 &       48.5 \\
VIVI-Ex(4)-AA           &     61.7 &     56.3 &        80.9 &       57.0 \\
VIVI-Ex(4)-Big          &     61.9 &     56.9 &        80.6 &       57.1 \\
Ex-YT-F-AA              &     61.9 &     56.1 &        81.7 &       57.2 \\
VIVI-Ex(4)-Big-AA       &     63.5 &     58.2 &        81.5 &       59.0 \\
Semi-Ex-10\%             &     63.9 &     69.3 &        80.3 &       50.9 \\
Sup-100\%                &     65.6 &     72.6 &        82.2 &       51.1 \\
VIVI-Ex(4)-Co(10\%)      &     66.3 &     63.1 &        82.4 &       61.1 \\
Sup-Rot-100\%            &     67.5 &     73.1 &        83.2 &       54.8 \\
VIVI-Ex(4)-Co(100\%)     &     69.2 &     70.1 &        83.2 &       61.3 \\
VIVI-Ex(4)-Co(100\%)-Big &     70.4 &     72.2 &        83.7 &       62.3 \\
\bottomrule
\end{tabular}
    \caption{Overall and per group mean testing accuracy for fine-tuning hyper-parameter selection according to {\bf Version 2 (arXiv:1910.04867v2) of the VTAB benchmark}. Each number is the median of three fine-tuning runs. See the caption of Table~\ref{tab:full-results-v2} for a description of the method abbreviations.}
    \label{tab:per-group-results-v2}
\end{table*}

\begin{table*}[h!]
    \centering
    \small
    \small
\newcommand{\green}[1]{\textcolor{Green}{(#1)}}
\begin{tabular}{lrlccc}
\toprule
\textsc{Method}                     & \multicolumn{2}{l}{\textsc{Mean}}                          & \textsc{Nat.} & \textsc{Spec.} & \textsc{Str.} \\
\midrule
Ex-ImageNet             &     59.2 & &       52.9 &       \bf 80.9 &       53.8 \\
VIVI-Ex(4)              &     61.0 &  \green{+1.8} &       55.6 &        79.7 &       56.4 \\
\color{lightgray} VIVI-Ex(2)              &     \color{lightgray} 61.4 &  \color{lightgray} (+2.2) &   \color{lightgray} 55.3 &        \color{lightgray} 79.9 &       \color{lightgray} 57.5 \\
VIVI-Ex(4)-Big          &     \bf 61.9 &  \green{+2.7}&      \bf 56.9 &        80.6 &      \bf 57.1 \\
          \midrule
Semi-Ex-10\%~\cite{zhai2019visual}             &     63.9 & &       \bf 69.3 &        80.3 &       50.9 \\
VIVI-Ex(4)-Co(10\%)      &     \bf 66.3 & \green{+2.4} &       63.1 &        \bf 82.4 &       \bf 61.1 \\
          \midrule
Sup-100\%                &     65.6 & &       72.6 &        82.2 &       51.1 \\
Sup-Rot-100\%~\cite{zhai2019visual}            &     67.5 & &       \bf 73.1 &        83.2 &       54.8 \\
VIVI-Ex(4)-Co(100\%)     &     69.2 & \green{+3.6} &     70.1 &        83.2 &       61.3 \\
VIVI-Ex(4)-Co(100\%)-Big &    \bf 70.4 & \green{+4.8} &       72.2 &        \bf 83.7 &       \bf 62.3 \\
\bottomrule
\end{tabular}
    \caption{Testing result summary as in Table~\ref{tab:cotraining} for fine-tuning hyper-parameter selection according to {\bf Version 2 (arXiv:1910.04867v2) of the VTAB benchmark}. Each number is the median of three fine-tuning runs.}
    \label{tab:cotraining-v2}
\end{table*}

\end{document}